\documentclass[journal,twoside]{IEEEtran}
\ifCLASSINFOpdf
\else
\fi
\usepackage{cite}
\usepackage{amsmath,amssymb,amsfonts}
\usepackage{algorithmic}
\usepackage{graphicx}
\usepackage{subfigure}
\usepackage{textcomp}
\usepackage{makecell}
\usepackage{color}

\begin{document}
%
\title{Map-based Visual-Inertial Localization:\\ Consistency and Complexity}


\author{\IEEEauthorblockN{Zhuqing Zhang\IEEEauthorrefmark{1},
Yanmei Jiao\IEEEauthorrefmark{1},
Shoudong Huang\IEEEauthorrefmark{2},
Yue Wang\IEEEauthorrefmark{1}, and
Rong Xiong\IEEEauthorrefmark{1}\\}
\IEEEauthorblockA{\IEEEauthorrefmark{1}State Key Laboratory of Industrial Control Technology, \\Zhejiang University, Hangzhou, Zhejiang, China.\\}
\IEEEauthorblockA{\IEEEauthorrefmark{2}Centre for Autonomous Systems, Faculty
of Engineering and Information Technology,\\ University of Technology
Sydney, Ultimo, NSW 2007, Australia}
\thanks{Corresponding author: Yue Wang (email: wangyue@iipc.zju.edu.cn)}}


%



\IEEEtitleabstractindextext{%
\begin{abstract}
Drift-free localization is essential for autonomous vehicles. In this paper, we address the problem by proposing a filter-based framework, which integrates the visual-inertial odometry and the measurements of the features in the pre-built map. In this framework, the transformation between the odometry frame and the map frame is augmented into the state and estimated on the fly. Besides, we maintain only the keyframe poses in the map and employ Schmidt extended Kalman filter to update the state partially, so that the uncertainty of the map information can be consistently considered with low computational cost. Moreover, \textcolor{black}{we theoretically demonstrate that the ever-changing linearization points of the estimated state can introduce spurious information to the augmented system and make the original four-dimensional unobservable subspace vanish,} leading to inconsistent estimation in practice. To relieve this problem, we employ first-estimate Jacobian (FEJ) to maintain the correct observability properties of the augmented system. Furthermore, we introduce an observability-constrained updating method to compensate for the significant accumulated error after the long-term absence (can be $\mathbf{3}$ minutes and $\mathbf{1}$ km) of map-based measurements. Through simulations, the consistent estimation of our proposed algorithm is validated. Through real-world experiments, we demonstrate that our proposed algorithm runs successfully on four kinds of datasets with the lower computational cost ($\mathbf{20} \boldsymbol{\%}$ time-saving) and the better estimation accuracy ($\mathbf{45} \boldsymbol{\%}$ trajectory error reduction) compared with the baseline algorithm VINS-Fusion, \textcolor{black}{whereas VINS-Fusion fails to give bounded localization performance on three of four datasets because of its inconsistent estimation.}
\end{abstract}

}

\maketitle

\IEEEdisplaynontitleabstractindextext

%
\IEEEpeerreviewmaketitle

\section{Introduction}
%
%
%
%
\IEEEPARstart{A}{ccurate} localization is primary for the functioning of an autonomous vehicle. \textcolor{black}{With equipped sensors (e.g. cameras) on the vehicle, the environment features can be extracted. By matching these features with the ones in a pre-built map, the pose of the vehicle in the map reference frame can be obtained. Then, the planned path in the pre-built map can be transformed into the frame of the vehicle by the pose of the vehicle, such that the vehicle can be guided from the current position to the destination.} Especially, information like routing and transportation infrastructure in the map can be utilized to improve the accuracy of the localization\cite{dxq,yh,SC,jym}.
\textcolor{black}{We consider that an ideal localization system should perform the error-bounded pose estimation, maintain the consistency so that the estimated covariance can be used for information fusion, and have the computational cost affordable by onboard devices with limited computational capability}.

In recent years, visual-inertial localization has become an attractive solution because of the mature and low-cost vision and inertial sensors.
\textcolor{black}{Most studies focus on consistent and lightweight visual-inertial odometry (VIO). However, these VIO systems will suffer from unbounded drift for long-term running, which is insufficient for autonomous vehicle applications \cite{msckf,openvins,vinsmono}.} To compensate for the drift, one way is to introduce the global positioning system (GPS) \cite{vinsgps,gomsf,gpsvio}. \textcolor{black}{Nevertheless}, GPS can sometimes be unstable especially in \textcolor{black}{the city canyon or parking lot environments}. Another widely used method is to employ a pre-built map, which can bound the error by matching the map \textcolor{black}{features} with the on-the-fly sensor reading \cite{getout,maplab,orbslam3}. However, less attention is paid \textcolor{black}{to} the theoretic analysis for such fusion of map-based measurements and visual-inertial odometry. In this paper, we focus on the \textcolor{black}{computational complexity} and consistency of the map-based visual-inertial localization.

\textcolor{black}{A typical visual map could contain hundreds of (say $m$) keyframes, and tens of thousands of (say $n$ and $m \ll n$) point features \cite{maplab}}. Besides, \textcolor{black}{the map information is actually imperfect and is characterized by uncertainty. The uncertainty is usually represented by a covariance matrix, which requires the storage with quadratic complexity.} To make the localization system run in real-time, different trade-offs have been made in the previous efforts. 

 \textcolor{black}{To integrate map information, an intuitive and theoretically sound way is keeping all the map-related variables (e.g. map features) in the state vector, and the observations of the map features are used to update the state in the manner of the standard extended Kalman filter (EKF), equivalent to the formulation of EKF simultaneous localization and mapping (EKF-SLAM).} However, since the computational complexity of updating covariance is $\mathcal{O}(n^2)$, \textcolor{black}{this makes the system difficult to run in real-time, especially for a large map.} To solve this problem, \cite{consistent} proposes a system based on Schmidt-EKF \cite{schmidtekf}, which discards the updating of the map state, leading to a conservative \textcolor{black}{yet} consistent estimation. The advantage of this operation is that it largely reduces the computation complexity ($\mathcal{O}(n)$). However, \textcolor{black}{the method in \cite{consistent}} still stores the whole map in the state vector, which makes the state dimension quite high and needs lots of memory ($\mathcal{O}(n^2)$) for the covariance. Another framework for efficient fusion is multi-sensor fusion \cite{msf}, which follows a loosely coupled fashion, i.e., the estimation process of the local odometry is independent of map-based information \cite{survey vio}. Besides, \cite{msf} regards multiple map feature measurements as an integrated measurement to update \textcolor{black}{the relative transformation between the local odometry reference frame and the map reference frame}. This mechanism is also applied in Maplab \cite{maplab}. As the map information will not be added into the state vector, the dimension of the state vector has a small constant value and can be neglected compared with the dimension of the map information $n$, which leads to the low computational complexity ($\mathcal{O}(1)$). \textcolor{black}{However, this kind of state formulation like \cite{msf} and \cite{maplab}} ignores the uncertainty of the map landmarks as well as the correlation between the current state and the map information, bringing overconfident estimation.

A general idea in the methods mentioned above is introducing a variable, i.e., the relative transformation between \textcolor{black}{the} odometry frame and \textcolor{black}{the} map frame, which is estimated on the fly. We call this variable as the augmented variable and the state including this augmented variable as the augmented state. Ideally, the augmented variable should be constant. However, in practice, due to the ever-changing estimated values of the augmented variable, the Jacobian matrix of the observation function derived from the estimated augmented variable will also change accordingly. \textcolor{black}{Therefore, spurious information is introduced into the system, bringing inconsistency, producing overconfident estimation, and finally resulting in inaccurate or even divergent results.} To the best of our knowledge, the observability properties of this augmented state have not been analyzed in the previous works and remain vague.


In this paper, we propose a filter-based drift-free localization framework with low \textcolor{black}{computational} complexity \textcolor{black}{yet} consistency. To be specific, we represent the map with $m$ keyframes and corresponding covariances, as well as $n$ landmarks (no covariance). \textcolor{black}{In this way, the required storage of covariance is $\mathcal{O}(m^2+n)$, much smaller than $\mathcal{O}(m^2+n^2)$ as $m \ll n$}. Then, we maintain the poses of the map keyframes in the state vector. When updating the state, we project the landmark-related Jacobian matrix to its null space so that we do not need to consider the covariances of the landmarks. To avoid the large computation of updating the map keyframes, whose computational complexity is $\mathcal{O}(m^2)$, we apply Schmidt-EKF for map state updating to reduce the computational complexity to $\mathcal{O}(m)$. Moreover, we theoretically demonstrate that for the ideal case, the augmented system has four dimensions of unobservable subspace, while for the real case, such unobservable subspace vanishes, resulting in inconsistency. The reason lies in the ever-changing linearization points, which make the filter gain spurious information in the state space along the directions where no information is actually available. To relieve this problem, we propose the first-estimate Jacobian (FEJ) to fix the linearization points in the augmented system. In addition, to correct large drift, we propose an error compensated updating step while keeping the linearization point still fixed, which can be useful when map-based measurements \textcolor{black}{are} available after a long duration of absence. Finally, we evaluate the proposed localization system using simulation data as well as datasets collected from three types of vehicles, showing the theory validness and system effectiveness. To summarize, the contributions of this paper are four-folded:
\begin{itemize}
	\item \textcolor{black}{Propose a map-based drift-free visual-inertial localization system that maintains consistent state estimation and low computational complexity.}
	\item Map uncertainty is taken into \textcolor{black}{consideration} while keeping the \textcolor{black}{low computational complexity with Schmidt-EKF}.
	\item Theoretical analyses are made on the observability of the augmented state formulation, which inspires the FEJ, and the error compensated updating techniques for better estimation.
	\item Substantial experiments on three types of vehicles as well as simulations are conducted to validate the theory and the performance of the full system.
\end{itemize}

\section{Related works}

\subsection{Filter-based visual-inertial localization}

One branch of the visual-inertial localization methods utilizes a filter for camera and IMU fusion. Early efforts solve the fusion problem in a loosely coupled mechanism. The IMU is modeled in the propagation process, and the relative motion calculated by the vision sensor is employed to update the state \cite{combine vi}. A more recent loosely coupled filter, multi-sensor fusion (MSF), introduces an augmented-state-based state formulation, which leverages the positional measurement from visual odometry for updating, and becomes a general option in later works \cite{msf}. \textcolor{black}{This solution is also employed on aerial vehicles for autonomous flight and mapping in \cite{automousvbf}.}

Compared with the loosely coupled filter, a more accurate solution is the tightly coupled filter. An early work following this mechanism estimates the poses and the map landmarks at the same time \cite{vi_nml}. Thus, the computation for such formulation grows exponentially with the number of landmarks. \textcolor{black}{To improve the computational efficiency,} popular multi-state constraint Kalman filter (MSCKF) is proposed to project the landmarks to the null space of their Jacobian matrices, resulting in a very fast and light-weighted implementation \cite{msckf}. To further eliminate the drift in visual-inertial odometry, GPS \cite{gpsvio} and map-based measurements \cite{consistent,maplab} are further introduced using the augmented state formulation for fusion. In these works, a trade-off has been made between the computation and the consideration of map uncertainty. On the contrary, in this paper, we design a filter \textcolor{black}{with the characteristics of consistency and low computational complexity at the same time}.

However, a major drawback of the filter is that the one-time linearization and marginalization will inevitably introduce errors\cite{vio_ar}. This error becomes more evident when the measurements are absent for a long time, which can be true for intermittent GPS or less feature matching. Nevertheless, this error is less discussed in drift-free visual-inertial localization systems.

\subsection{Optimization based visual-inertial localization}

Another branch of visual-inertial localization is based on graph optimization. In this formulation, the measurements are formalized as a graph, which is optimized by iterative algorithms. Following this idea, popular visual odometry methods like \cite{vinsmono, orbslam3} are proposed and utilized in many aerial and ground moving vehicles. To reduce the optimization variables, preintegration is proposed as a factor between the two keyframes \cite{preintegration}. Both methods employ a windowed optimization for a trade-off between the accuracy and the computation.

To further reduce the drift of the visual-inertial odometry, global measurements can also be integrated into the optimization-based visual-inertial localization. In VINS-Fusion\cite{vinsmono,vinsgps,vinslocal,vinscalib}, map-based measurements are fused in a outer-loop pose-graph optimization. In \cite{tcgp}, a tightly coupled approach is proposed to optimize all measurements together with respect to all states. When both odometry and map-based measurements are integrated, a degeneracy analysis is performed in \cite{degeneration}. In general, the optimization-based visual-inertial localization has superior accuracy than the filtering-based methods because the nonlinear system can be re-linearized with the cost of more expensive resources. The uncertainties of the map are ignored in all these methods.

\subsection{Observability analysis of visual-inertial system}

A typical drawback for all filter-based solutions is that the linearization error can cause inconsistency, resulting in overconfident covariance. \textcolor{black}{Early works demonstrate that the dimension of the unobservable subspace of the visual-inertial odometry is four \cite{obSFM}, three of which form the unobservable subspace along the translation direction and one of which forms the unobservable subspace along the yaw direction.} However, due to the linearization point of the landmark changes during the process, spurious information occurs in the unobservable directions, seriously breaking the consistency \cite{consistVIN} of the system. Several works try to relieve the inconsistency in visual-inertial odometry by fixing the linearization point \cite{fej} or searching for a linearization point in the observable space \cite{oc}. Unfortunately, since map-based measurements are introduced for updating via the augmented state formulation, observability analyses of visual-inertial odometry cannot reflect the observability of the augmented state. In \cite{gpsvio}, the observability of visual-inertial odometry fusing GPS is analyzed, which shows the insights for the observability of the augmented system. \textcolor{black}{However, \cite{gpsvio} only points out that the augmented system suffers from inconsistency and proposes a consistent estimation algorithm that is only suitable for gravity-aligned maps, i.e., the map has 4 degrees of freedom (DoF). To be specific, after initializing the relative transformation between the VIO frame and the 4 DoF map frame, the variables in the state vector are transformed from the 4 DoF VIO frame to the 4 DoF map frame, and the state vector is propagated by IMU measurements in the 4 DoF map frame. However, for a more general gravity-unaligned map (a 6 DoF map), the state cannot be correctly propagated by the IMU measurements in the map frame because the gravity in the map frame is unknown. On the contrary, in this paper, we propose a general framework that can be used for the case of 6 DoF maps.}






\section{Problem Statement}

We begin the statement of the drift-free visual-inertial localization system by introducing the frames in the problem.

\textbf{Frame definition} There are five frames in the visual-inertial localization problem, as shown in Fig. \ref{fig:coordinate}. The odometry origin inertial frame ${L}$, the current IMU body frame ${I}$, and the camera frame ${C}$, which captures all the information for the visual-inertial odometry. \textcolor{black}{The IMU outputs specific force and angular velocity in the frame $I$. By gravity compensation, we can obtain the acceleration measurements, which are accompanied by the angular velocity measurements as the input of the visual-inertial localization system.} And the camera observes landmarks in ${C}$. \textcolor{black}{The transformation between ${I}$ and ${C}$ is assumed known}. The map origin frame ${G}$ and the map keyframe ${KF}$ record the map information. Note that ${G}$ may not be an inertial reference frame. The map landmarks are represented in ${KF}$, which are all defined in ${G}$. \textcolor{black}{The poses of map keyframes in $G$ and the positions of all map landmarks anchored in ${KF}$ are determined when the map is built.} When there are matchings between the current view and the map features, we have a map-based measurement bridging the frame ${G}$, ${KF}$ and ${C}$, which can ease the drift in visual-inertial odometry.

\textbf{Drift-free visual-inertial localization} Based on the frames defined above, \textcolor{black}{we know there are two unknown transforms, one is the transform between ${G}$ and ${L}$, and the other is the transform between ${L}$ and ${I}$. Visual-inertial odometry yields the estimation of the latter with drift. The problem of a drift-free visual-inertial localization system is then defined as estimating the former and correcting the latter, so that the transformation between ${G}$ and ${I}$ is drift-free by leveraging the local feature tracking and map-based measurements.} \textcolor{black}{The design principles are consistency and low computational complexity.}

\section{System Modeling and Filter Design}
\label{system}

The system model consists of two parts. The visual-inertial odometry part and the map-based measurements part. MSCKF-style filtering \cite{msckf} is applied for visual-inertial odometry, while Schmidt filtering \cite{schmidtekf} is applied for the map-based measurement updating.


\subsection{Visual-Inertial Odometry}

The state vector at time step $k$ is defined as
\begin{equation}
    \mathbf{x}_k= \left[\begin{array}{cc}
        \mathbf{x}_{I_k}^{\top} & \mathbf{x}_{C_k}^{\top}  \\ \end{array} \right]^{\top},
\end{equation}
where $\mathbf{x}_{I_k}$ is the current state of the IMU, the transform between $I$ and $L$. We further define $\mathbf{x}_{I_k}$ as
\begin{equation} \label{eq:imu_state}
    \mathbf{x}_I=\left[\begin{array}{ccccc} {}^{I_k}\mathbf{q}_{L}^{\top}&  {}^{L}\mathbf{v}_{I_k}^{\top} &{}^{L}\mathbf{p}_{I_k}^{\top} & \mathbf{b}_{g_k}^{\top}& \mathbf{b}_{a_k}^{\top}\\ \end{array} \right]^{\top},
\end{equation}
where ${}^{I_k}\mathbf{q}_{L}$ is an unit quaternion variable, defined by JPL\cite{jpl}. Its corresponding rotation matrix ${}^{I_k}\mathbf{R}_L$ rotates a 3D vector from the odometry inertial frame ${L}$ into the IMU frame ${I}$. ${}^{L}\mathbf{p}_{I_k}$ is the position of the body in ${L}$ at time step $k$. ${}^{L}\mathbf{v}_{I_k}$ is the velocity of the body in frame ${L}$. $\mathbf{b}_{g_k}$ and $\mathbf{b}_{a_k}$ are the gyroscope and accelerometer bias. The clone state $\mathbf{x}_{C_k}$ is defined as
\begin{equation}
\begin{aligned}
  \mathbf{x}_{C_k}&= \left[\begin{matrix} {}^{I_{k-1}}\mathbf{q}_{L}^{\top}& {}^{L}\mathbf{p}_{I_{k-1}}^{\top}&\dots &{}^{I_{k-N}}\mathbf{q}_{L}^{\top}& {}^{L}\mathbf{p}_{I_{k-N}}^{\top}\\ \end{matrix} \right]^{\top}\\
  &\triangleq \left[\begin{matrix}\mathbf{x}_{C_k,1}^{\top}&\dots&\mathbf{x}_{C_k,N}^{\top}\end{matrix} \right]^{\top}
\end{aligned}
\end{equation}
which consists of the clones of the $N$ latest historical body poses. $\mathbf{x}_{C_k}$ is the so-called sliding window. For each body pose, the corresponding camera pose can be obtained through the extrinsic between the body and the camera. For simplicity, in the following of the paper, we assume the extrinsic between the body and camera is the \textcolor{black}{identity} matrix, \textcolor{black}{i.e., the pose of the body is identical to the pose of the equipped camera.}

\textbf{State propagation} When we receive IMU data at time step $k$, the IMU kinematics model is used to propagate the body state from time step $k$ to $k+1$:

\begin{equation}\label{eq:state_prop}
\textcolor{black}{    \mathbf{x}_{k+1|k}=\mathbf{f}(\mathbf{x}_{k},\mathbf{a}_{m_k},\mathbf{n}_{a_k},\boldsymbol{\omega}_{m_k},\mathbf{n}_{g_k}),
}
\end{equation}
where $\mathbf{f}$ represents the IMU kinematics equations, $\mathbf{a}_{m_k}$ and $\mathbf{\omega}_{m_k}$ are the measurements (\textcolor{black}{body acceleration} and angular velocity) derived from the IMU. $\mathbf{n}_{a_k}$ and $\mathbf{n}_{g_k}$ are the zero-mean Gaussian white noise of the IMU measurements. With this function, we can also propagate the state covariance by
\begin{equation}\label{eq:prop}
    \mathbf{P}_{k+1|k}=\mathbf{\Phi}_{k}\mathbf{P}_{k}\mathbf{\Phi}_{k}^{\top}+\mathbf{G}_{k}\mathbf{Q}_{k}\mathbf{G}_{k}^{\top},
\end{equation}
where the subscript $k+1|k$ means the propagated value from time step $k$ to $k+1$.  $\mathbf{P}_{k}$ and $\mathbf{Q}_{k}$ are the system state covariance and the noise covariance, respectively. $\mathbf{\Phi}_{k}$ and $\mathbf{G}_{k}$ are the Jacobians of (\ref{eq:state_prop}) with respect to the system state and the noise, respectively. For the detailed derivation, please refer to \cite{msckf, openvins}.

\textbf{Observation function} By tracking the visual features in the images stream, at time step $k$, the cloned poses $\mathbf{x}_{C_k}$ can observe a set of features $\left\{f\right\}$. For each feature $f_i$ observed by $\mathbf{x}_{C_k,j},\, j=1\dots N$, we have the following observation function\footnote{For the sake of simplicity, in the following expressions of this section, we neglect the subscript $k$, so that $\mathbf{x}_{C_k,j}$ becomes $\mathbf{x}_{C,j}$. However, keep in mind that all the variables in the equations are at the time step $k$}:  
\textcolor{black}{\begin{equation}
    \mathbf{z}_{f_{ij}}=\mathbf{h}\left(\mathbf{x}_{C,j},\mathbf{f}_i\right)+ \mathbf{n}_{f_{ij}},
\end{equation}
where $\mathbf{z}_{f_{ij}}$ is the 2D measurement of $f_i$ in the image captured by the cloned pose $\mathbf{x}_{C,j}$. $\mathbf{f}_{i}$ is the 3D coordinate of $f_i$.} $\mathbf{n}_{f_{ij}}$ is the observation noise of the cloned camera $j$ (with the cloned pose $\mathbf{x}_{C,j}$) towards the feature $f_i$. By applying the first-order \textcolor{black}{Taylor series expansion} to the observation function at the current estimated state, we have:
\begin{equation}\label{eq:local_ob}
\textcolor{black}{
\begin{aligned}
    \mathbf{z}_{f_{ij}}& \approx \mathbf{h}\left(\hat{\mathbf{x}}_{C,j}, \hat{\mathbf{f}}_{i} \right) + {}_{i}\mathbf{H}_{\hat{\mathbf{x}}_{C,j}}\tilde{\mathbf{x}}_{C,j}+\mathbf{H}_{\hat{\mathbf{f}}_i,j}\tilde{\mathbf{f}}_i+\mathbf{n}_{f_{ij}},\\
    \mathbf{r}_{f_{ij}}&:=\tilde{\mathbf{z}}_{f_{ij}}= {}_{i}\mathbf{H}_{\hat{\mathbf{x}}_{C,j}}\tilde{\mathbf{x}}_{C,j}+\mathbf{H}_{\hat{\mathbf{f}}_i,j}\tilde{\mathbf{f}}_i+\mathbf{n}_{f_{ij}},
\end{aligned}}
\end{equation}
where $\hat{\cdot}$ represents the current estimated values and $\tilde{\cdot}$ represents the errors between the true values and the estimated values. \textcolor{black}{${}_{i}\mathbf{H}_{\hat{\mathbf{x}}_{C,j}}$ and $\mathbf{H}_{\hat{\mathbf{f}}_i,j}$ are the Jacobians of $\mathbf{h}$ linearized at the estimated cloned camera state $\hat{\mathbf{x}}_{C,j}$ and the observed feature $\hat{\mathbf{f}_i}$, respectively. The left-subscript $i$ of ${}_{i}\mathbf{H}_{\hat{\mathbf{x}}_{C,j}}$ means this Jacobian is related with $f_i$.}

\textcolor{black}{As we do not maintain features in the state vector, the term related with $\tilde{\mathbf{f}}_i$ should be marginalized. Thanks to the multi-state constraint, for each feature $\mathbf{f}_i$, it can be observed by multiple cloned cameras (say $\mathcal{M}$ cloned cameras). For each observation, we can get a linearized observation function by (\ref{eq:local_ob}). Stacking these $\mathcal{M}$ functions together, we have 
\begin{equation}\label{eq:local_error}
    \mathbf{r}_{f_i}= {}_{i}\mathbf{H}_{\hat{\mathbf{x}}_k}\tilde{\mathbf{x}}_k+\mathbf{H}_{\hat{\mathbf{f}}_i}\tilde{\mathbf{f}}_i+\mathbf{n}_{f_i},
\end{equation}
where $\mathbf{r}_{f_i}$, ${}_{i}\mathbf{H}_{\hat{\mathbf{x}}_k}$, $\mathbf{H}_{\hat{\mathbf{f}}_i}$ and $\mathbf{n}_{f_i}$ are stacked with the $\mathcal{M}$ $\mathbf{r}_{f_{ij}}$, ${}_{i}\mathbf{H}_{\hat{\mathbf{x}}_{C,j}}$, $\mathbf{H}_{\hat{\mathbf{f}}_i,j}$ and $\mathbf{n}_{f_{ij}}$, respectively. We can find that for $\mathcal{M}>1$, $\mathbf{H}_{\hat{\mathbf{f}}_i}$ has more rows than columns ($2\mathcal{M}>3$), which guarantees that $\mathbf{H}_{\hat{\mathbf{f}}_i}$ has the left null space $\mathbf{N}$, such that}
\begin{equation}\label{eq:local_ns}
    \mathbf{N}^{\top} \mathbf{r}_{f_i}=\mathbf{N}^{\top}
    {}_{i}\mathbf{H}_{\hat{\mathbf{x}}_k}\tilde{\mathbf{x}}_k+\mathbf{N}^{\top}\mathbf{H}_{\hat{\mathbf{f}}_i}\tilde{\mathbf{f}}_i+\mathbf{N}^{\top}\mathbf{n}_{f_i},
\end{equation}
\vspace{-0.6cm}
\begin{equation}\label{eq:local_final}
    \mathbf{r}^{\prime}_{f_i}=
{}_{i}\mathbf{H}^{\prime}_{\hat{\mathbf{x}}_k}\tilde{\mathbf{x}}_k+\mathbf{n}^{\prime}_{f_i},
\end{equation}
where $\mathbf{r}^{\prime}_{f_i}= \mathbf{N}^{\top} \mathbf{r}_{f_i}$, ${}_{i}\mathbf{H}^{\prime}_{\hat{\mathbf{x}}_k}=\mathbf{N}^{\top}
{}_{i}\mathbf{H}_{\hat{\mathbf{x}}_k}$, $\mathbf{n}^{\prime}_{f_i}=\mathbf{N}^{\top}\mathbf{n}_{f_i}$. We would employ ${}_{i}\mathbf{H}^{\prime}_{\hat{\mathbf{x}}_k}$ and $\mathbf{r}^{\prime}_{f_i}$ to update the state.

\subsection{Map-based Measurement}
\label{global model}


\begin{figure}[t!]
    \centering
    \setlength{\abovecaptionskip}{0cm}
    \includegraphics[width=1\linewidth]{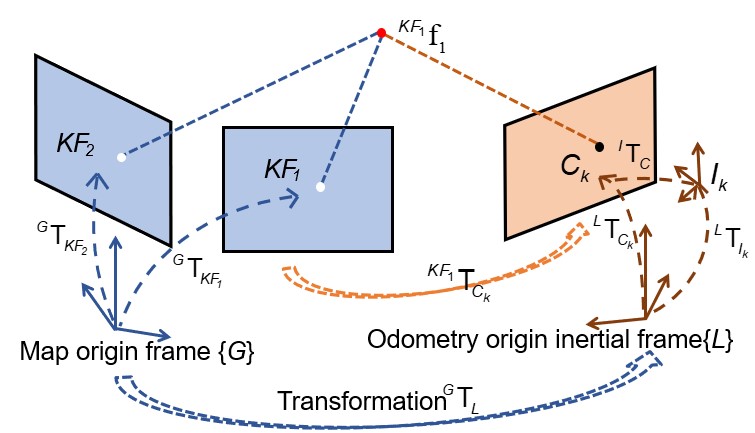}
     \caption{The global observation function}
     \label{fig:coordinate}
     \vspace{-0.6cm}
\end{figure}

\textcolor{black}{This part is the key augmentation to the VIO system. With the map-based measurement illustrated below, the map uncertainty can be incorporated into the system.} Given a pre-built map, we have feature matchings between the current image (${C}_k$ in Fig. \ref{fig:coordinate}) and keyframes in the map (e.g. ${KF_1}$ and ${KF_2}$ in Fig. \ref{fig:coordinate}). Different from tracked features from the online image stream, the matched map features have 3D positions (the red dot in Fig. \ref{fig:coordinate}). We represent the 3D features in the coordinate system of a map keyframe observing it, called the anchored keyframe. This measurement provides a bridge between the map frame and the visual-inertial odometry frame to \textcolor{black}{ease the drift of the odometry.}

\textbf{Augmented state formulation} By applying the techniques of 3D-2D pose estimation e.g., EPnP\cite{epnp}, we have the relative pose ${}^{{KF_1}}\mathbf{T}_{{C}_k}$ between the keyframe and the current frame. This value, combined with the matched keyframe pose ${}^{G}\mathbf{T}_{{KF_1}}$ and the current frame pose ${}^{L}\mathbf{T}_{{C}_k}$, will be used to initialize ${}^{G}\mathbf{T}_{L}$. After the initialization, $\mathbf{x}_T \triangleq {}^{G}\mathbf{T}_{L}$ will be augmented into the state vector:
\begin{equation}
    \mathbf{x}_k= \left[\begin{array}{ccc}
        \mathbf{x}_{I_k}^{\top} & \mathbf{x}_{C_k}^{\top} & \mathbf{x}_{T_k}^{\top}  \\ \end{array} \right]^{\top}.
\end{equation}
Besides, the covariance of the state would also be augmented. As the initial estimation of $\mathbf{x}_T$ may be inaccurate, its covariance can be assigned with a big value. It should be noted that every time we have matches from the map, the keyframes' poses are added to the state vector:
\begin{equation} \label{eq:full state}
    \mathbf{x}_k= \left[\begin{array}{cccc}
        \mathbf{x}_{I_k}^{\top} & \mathbf{x}_{C_k}^{\top} & \mathbf{x}_{T_k}^{\top} & \mathbf{x}_{KF}^{\top} \\ \end{array} \right]^{\top},
\end{equation}
\begin{equation}
\begin{aligned}
    \mathbf{x}_{KF}&= \left[\begin{matrix}
        {}^{G}\mathbf{q}_{KF_1}^{\top}& {}^{G}\mathbf{p}_{KF_1}^{\top}&\dots &{}^{G}\mathbf{q}_{KF_M}^{\top}& {}^{G}\mathbf{p}_{KF_M}^{\top} \\ \end{matrix} \right]^{\top}\\
        &\triangleq \left[\begin{matrix}
        \mathbf{x}_{KF_1}^{\top}&\dots \mathbf{x}_{KF_M}^{\top} \\ \end{matrix} \right]^{\top},\\
\end{aligned}
\end{equation}
where $\mathbf{x}_{KF_i}$ is the pose of keyframe $KF_i$, i.e., $\mathbf{x}_{KF_i}\triangleq {}^{G}\mathbf{T}_{KF_i}$.

With this augmented state system, the map-based measurements can be modeled and utilized to correct \textcolor{black}{the drift of the pure VIO}.

For briefness, we divide the state into two parts: the active part $\mathbf{x}_A\triangleq \left[\begin{matrix}
        \mathbf{x}_I^{\top} & \mathbf{x}_C^{\top} & \mathbf{x}_T^{\top}\\ \end{matrix} \right]^{\top}$, and the nuisance part $\mathbf{x}_N\triangleq \mathbf{x}_{KF}$, leading to the following expressions:
\textcolor{black}{
\begin{equation}\label{eq:xaxn}
\begin{aligned}
     \mathbf{x}_k&= \left[\begin{array}{cc}
        \mathbf{x}^{\top}_{A_k} & \mathbf{x}_{N_k}^{\top}\\ \end{array} \right]^{\top},\\
    \mathbf{P}_{k}& =\begin{bmatrix}\mathbf{P}_{AA_k}&\mathbf{P}_{AN_k}\\
        \mathbf{P}_{AN_k}^{\top}&\mathbf{P}_{NN_K}\end{bmatrix},
\end{aligned}
\end{equation}}
where $\mathbf{P}_{AA_k}$ and $\mathbf{P}_{NN_K}$ are the self-covariance of $\mathbf{x}_{A_k}$ and $\mathbf{x}_{N_k}$, respectively, and $\mathbf{P}_{AN_k}$ is the cross-covariance between $\mathbf{x}_{A_k}$ and $\mathbf{x}_{N_k}$. This kind of partition is very useful for the Schmidt-EKF updating, which is introduced in Sec. \ref{algorithm}.

\textbf{Single frame observation function} As is shown in Fig. \ref{fig:coordinate}, supposing there are feature matchings between the current image and multiple keyframes (for example, ${KF_1}$ and ${KF_2}$) at time step $k$, for each 3D map feature (for example, $^{{KF}_1}\mathbf{f}_{1}$), we can project it from the anchored keyframe (${KF}_1$) coordinate system into the 2D current image frame by the nonlinear observation function $^{1}\mathbf{g}$:
\begin{equation}\label{eq:global_observation_function}
    {}^{1}\mathbf{z}_{f}={}^{1}\mathbf{g}(\mathbf{x}_{KF_1},\mathbf{x}_{A},^{{KF}_1}\mathbf{f}_{1})+ {}^{1}\mathbf{n}_f,
\end{equation}
where the subscript $k$ is omitted for simplicity, whereas all the variables in the equations are at the time step $k$. ${}^{1}\mathbf{z}_{f}$ is a 2D pixel coordinate in the current image. \textcolor{black}{${}^{1}\mathbf{n}_f$ is the observation noise of the camera towards the 3D map feature (for example, $^{{KF}_1}\mathbf{f}_{1}$).} Linearizing the above function at the estimated values, like (\ref{eq:local_ob}), we have
\begin{equation}\label{eq:global_ob}
    {}^{1}\mathbf{r}_{f}={}^{1}\mathbf{H}_{\hat{\mathbf{x}}_{{A}_k}}\tilde{\mathbf{x}}_{{A}_k}+{}^{1}\mathbf{H}_{\hat{\mathbf{x}}_{{N}_k}}\tilde{\mathbf{x}}_{{N}_k}+{}^{1}\mathbf{H}_{^{{KF}_1}\hat{\mathbf{f}}_{1}} {}^{{KF}_1}\tilde{\mathbf{f}}_{1}+{}^{1}\mathbf{n}_f.
\end{equation}
The subscripts ${A}$ and ${N}$ represent the active part and the nuisance part of the state vector, respectively (cf. (\ref{eq:xaxn})).

The 3D map feature ${}^{{KF}_1}\mathbf{f}_{1}$ can also be observed by the anchored keyframe. Therefore the map feature can be reprojected into the anchored keyframe ${KF}_1$ by the reprojection function $^{2}\mathbf{g}$:
\begin{equation}
    \label{eq:project1}
    {}^{2}\mathbf{z}_{f}={}^{2}\mathbf{g}({}^{{KF}_1}\mathbf{f}_{1})+{}^{2}\mathbf{n}_f,
\end{equation}
\begin{equation}
    \label{eq:project1_error}
    {}^{2}\mathbf{r}_{f}={}^{2}\mathbf{H}_{^{{KF}_1}\hat{\mathbf{f}}_{1}} {}^{{KF}_1}\tilde{\mathbf{f}}_{1}+{}^{2}\mathbf{n}_f,
\end{equation}
\textcolor{black}{where ${}^{2}\mathbf{z}_{f}$ is the 2D pixel coordinate in $KF_1$, and ${}^{2}\mathbf{n}_f$ is the observation noise of the camera towards the 3D map feature when building the map.}

\textbf{Multi-frame observation function} Note that the 3D landmark ${}^{{KF}_1}\mathbf{f}_{1}$ can also be observed by the other map keyframe(s), like ${KF}_2$ in Fig. \ref{fig:coordinate}. Thus we have another reprojection function ${}^{3}\mathbf{g}$ which reprojects ${}^{{KF}_1}\mathbf{f}_{1}$ into the ${KF}_2$ image plane:
\begin{equation}
    \label{eq:project2}
    {}^{3}\mathbf{z}_{f}={}^{3}\mathbf{g}(\mathbf{x}_{{KF}_1},\mathbf{x}_{{KF}_2},{}^{{KF}_1}\mathbf{f}_{1})+{}^{3}\mathbf{n}_f,
\end{equation}
\begin{equation}
    \label{eq:project2_error}
    {}^{3}\mathbf{r}_{f}={}^{3}\mathbf{H}_{\hat{\mathbf{x}}_{{N}_k}}\tilde{\mathbf{x}}_{{N}_k}+{}^{3}\mathbf{H}_{^{{KF}_1}\hat{\mathbf{f}}_{1}} {}^{{KF}_1}\tilde{\mathbf{f}}_{1}+{}^{3}\mathbf{n}_f,
\end{equation}
\textcolor{black}{where ${}^{3}\mathbf{z}_{f}$ is the 2D pixel coordinate in $KF_2$, and ${}^{3}\mathbf{n}_f$ is the observation noise of the camera towards the 3D map feature when building the map, which is usually identical to ${}^{2}\mathbf{n}_{f}$.}
By stacking (\ref{eq:global_ob}), (\ref{eq:project1_error}) and (\ref{eq:project2_error}), we could get a 6-dimension measurement for the 3D landmark ${{}^{KF_1}\mathbf{f}}_{1}$.

In summary, whenever there are feature matchings between the current image and a single keyframe or multiple keyframes, for each 3D landmark ${}^{{KF}_j}\mathbf{f}_{i}$ anchored in the keyframe ${KF}_j$, \textcolor{black}{we can formulate the observation functions by (\ref{eq:global_ob}), (\ref{eq:project1_error}) and (\ref{eq:project2_error}). Stacking these functions together,} we have the observation function with the following form:
\begin{equation}\label{eq:global ob stack}
\textcolor{black}{
    {}^{M}\mathbf{r}_{f_i}= {}_{i}\mathbf{H}_{\hat{\mathbf{x}}_{{A}_k}}\tilde{\mathbf{x}}_{{A}_k}+{}_{i}\mathbf{H}_{\hat{\mathbf{x}}_{{N}_k}}\tilde{\mathbf{x}}_{{N}_k}+\mathbf{H}_{^{{KF}_j}\hat{\mathbf{f}}_{i}} {}^{{KF}_j}\tilde{\mathbf{f}}_{i}+{}^{M}\mathbf{n}_{f_i},
    }
\end{equation}
\textcolor{black}{where ${}^{M}\mathbf{r}_{f_i}$ is the stack of ${}^{1}\mathbf{r}_{f},{}^{2}\mathbf{r}_{f}$ (and ${}^{3}\mathbf{r}_{f}$), and we can get ${}_{i}\mathbf{H}_{\hat{\mathbf{x}}_{{A}_k}}$, ${}_{i}\mathbf{H}_{\hat{\mathbf{x}}_{{N}_k}}$, $\mathbf{H}_{^{{KF}_j}\hat{\mathbf{f}}_{i}}$ and ${}^{M}\mathbf{n}_{f_i}$ in a similar way. The left-superscript $M$ means the related values are derived from the \textbf{M}ap-based measurements and is used to distinguish the map-based measurements from the local VIO-based measurements (cf. (\ref{eq:local_error})).}

\textbf{Null space projection} It is important to note that since we regard map features as variables to consider map uncertainties, the Jacobian matrix with respect to map features $\mathbf{H}_{^{{KF}_j}\hat{\mathbf{p}}_{f_i}}$ needs to be computed. Otherwise, the landmarks would be regarded as constants, like \cite{maplab}, which leaves the system inconsistent. However, different from \cite{consistent}, we do not maintain map landmarks in the state vector, so we need to marginalize ${}^{{KF}_j}\mathbf{p}_{f_i}$. For each landmark, it has at least two measurements (\textcolor{black}{in} the current frame and the matched one or more keyframes). Note that in practice, the pose of the current frame can hardly be identical to that of the map frame. Therefore, the Jacobian matrices derived from these measurements are independent, leading to the row number of $\mathbf{H}_{^{{KF}_j}\hat{\mathbf{p}}_{f_i}}$ (being equal or greater than four) is more than its column number (in our case is three). This fact guarantees that we can find a left null space of  $\mathbf{H}_{^{{KF}_j}\hat{\mathbf{p}}_{f_i}}$ to marginalize the item related to the feature, as analyzed in  (\ref{eq:local_error})-(\ref{eq:local_final}). In this way, by multiplying both sides of (\ref{eq:global ob stack}) with the left null space of $\mathbf{H}_{^{{KF}_j}\hat{\mathbf{p}}_{f_i}}$, we have:
\textcolor{black}{
\begin{equation}\label{eq:global_error}
     \mathbf{r}_{f_i}^{*}={}_{i}\mathbf{H}^{*}_{\hat{\mathbf{x}}_{{A}_k}}\tilde{\mathbf{x}}_{{A}_k}+{}_{i}\mathbf{H}^{*}_{\hat{\mathbf{x}}_{{N}_k}}\tilde{\mathbf{x}}_{{N}_k}+\mathbf{n}_{f_i}^{*}.
\end{equation}}
Note that (\ref{eq:global_error}) only considers one landmark. We need to stack (\ref{eq:global_error}) for all matched landmarks to get the final observation function. By stacking multiple ${}_{i}\mathbf{H}^{*}_{\hat{\mathbf{x}}_{{A}_k}}$, ${}_{i}\mathbf{H}^{*}_{\hat{\mathbf{x}}_{{N}_k}}$ and $\mathbf{n}_{f_i}^{*}$, we get $\mathbf{H}^{*}_{{{A}_k}}$, $\mathbf{H}^{*}_{{{N}_k}}$ and $\mathbf{n}_{k}^{*}$, which leads to the following expression:
\begin{equation}\label{eq:global schmidt}
    \begin{aligned}
     \mathbf{r}^{*}_{k}&=\left[\begin{array}{cc}
     \mathbf{H}^{*}_{A_k} & \mathbf{H}^{*}_{N_k} \\
 \end{array}\right]
 \left[\begin{array}{c}
      \tilde{\mathbf{x}}_{A_k}  \\
      \tilde{\mathbf{x}}_{N_k}
 \end{array}\right] + \mathbf{n}^{*}_{k}.\\
    \end{aligned}
\end{equation}
Note that $\mathbf{H}^{*}_{N_k}$ represents the Jacobians of the related poses of the matched map keyframes; $\mathbf{H}^{*}_{A_k}$ represents the Jacobians of the current state $\mathbf{x}_{I_k}$ and the relative transform $\mathbf{x}_{T_k}$. Considering the state partition of (\ref{eq:xaxn}), we have the following concise expression:
\begin{equation}\label{eq:global_final}
     \mathbf{r}^{*}_{k}=\mathbf{H}_{k}^{*}\tilde{\mathbf{x}}_{k}+\mathbf{n}^{*}_k,
\end{equation}
which is used to perform the map-based updating.

\subsection{State Updating}
\label{algorithm}

Based on the system modeling introduced above, we can demonstrate how to update the state with both online tracked features (local features) and map-based measurements.

\textbf{Standard EKF update for odometry} As indicated by (\ref{eq:xaxn}) and (\ref{eq:global schmidt}), the state vector can be divided into the active part and the nuisance part. The online tracked features are used to update the state of the active part, i.e., odometry. This procedure is not related to the nuisance part, and  (\ref{eq:global schmidt}) is simplified as
\begin{equation}\label{eq:active part}
\begin{aligned}
 \mathbf{r}^{*}_{k}&=\mathbf{H}^{*}_{A_k}\tilde{\mathbf{x}}_{A_k}+\mathbf{n}^{*}_{k},
\end{aligned}
\end{equation}
after which, Kalman gain is obtained via
\textcolor{black}{
\begin{equation}
    \mathbf{K}_{A_k}=\mathbf{P}_{{AA}_{k|k-1}}\mathbf{H}^{*\top}_{A_k}\mathbf{S}^{-1}_{k},
\end{equation}}
and the state and covariance update equations for EKF are as follows:
\begin{equation}\label{eq:state_update}
\textcolor{black}{
    \hat{\mathbf{x}}_{A_k}=\hat{\mathbf{x}}_{A_{k|k-1}} + \mathbf{K}_{A_k}\mathbf{r}^{*}_k,
    }
\end{equation}

\begin{equation}\label{eq:update_cov}
\textcolor{black}{
    \mathbf{P}_{AA_{k}}=\mathbf{P}_{AA_{k|k-1}}-\mathbf{K}_{A}\mathbf{H}^{*}_{k}\mathbf{P}_{AA_{k|k-1}},
    }
\end{equation}
where $\hat{\mathbf{x}}_{A_{k|k-1}}$ and $\mathbf{P}_{AA_{k|k-1}}$ are generated \textcolor{black}{by} the propagation step (cf. (\ref{eq:state_prop}), (\ref{eq:prop})), and
\begin{equation}\label{S}
\textcolor{black}{
\mathbf{S}_k=\mathbf{H}_{A_k}^{*}\mathbf{P}_{AA_{k|k-1}}\mathbf{H}^{*\top}_{A_k}+\mathbf{R}_k,
}
\end{equation}
\textcolor{black}{where $\mathbf{R}_k$ is the covariance of observation noise at time step $k$.}

\textbf{Limitation of standard EKF update} When map-based measurements come, we have the standard Kalman gain as:
\textcolor{black}{
\begin{equation}\label{eq:expand_k}
\begin{aligned}
    \mathbf{K}_{k}=\left[\begin{matrix}\mathbf{K}_{A_k}\\\mathbf{K}_{N_k}\end{matrix}\right] &= \left[\begin{matrix}\mathbf{P}_{AA_{k|k-1}}\mathbf{H}_{A_k}^{*\top} + \mathbf{P}_{AN_{k|k-1}}\mathbf{H}_{N_k}^{*\top}\\\mathbf{P}_{NA_{k|k-1}}\mathbf{H}_{A_k}^{*\top} + \mathbf{P}_{NN_{k|k-1}}\mathbf{H}_{N_k}^{*\top}\end{matrix}\right]\mathbf{S}_{k}^{-1}\\
    &\triangleq\left[\begin{matrix}\bar{\mathbf{K}}_{A_k}\\\bar{\mathbf{K}}_{N_k}\end{matrix}\right]\mathbf{S}_{k}^{-1},
    \end{aligned}
\end{equation}}
and the \textcolor{black}{covariance is updated by}:

\begin{equation}
\textcolor{black}{
    \label{eq:expand_update_cov}
    \begin{aligned}
      \mathbf{P}_{k}&=\mathbf{P}_{k|k-1}-\\
      &\left[\begin{matrix} \mathbf{K}_{A_k}\mathbf{S}_k\mathbf{K}_{A_k}^{\top}& \mathbf{K}_{A_k}\mathbf{H}_{k}^{*} \left[\begin{matrix} \mathbf{P}_{AN_{k|k-1}}\\\mathbf{P}_{NN_{k|k-1}}
    \end{matrix}\right]\\
    \left[\begin{matrix} \mathbf{P}_{AN_{k|k-1}}\\\mathbf{P}_{NN_{k|k-1}}\\
    \end{matrix}\right]^{\top} \mathbf{H}_{k}^{*\top} \mathbf{K}_{A_k}^{\top}& \mathbf{K}_{N_k}\mathbf{S}_{k}\mathbf{K}_{N_k}^{\top}\end{matrix}\right].
    \end{aligned}
    }
\end{equation}

\textcolor{black}{From (\ref{eq:expand_update_cov}), one can see that the computation is dominated by the nuisance variables, i.e., the map information. To be specific, the heaviest computation is $\mathbf{K}_{N_k}\mathbf{S}_{k}\mathbf{K}_{N_k}^{\top}$, which is quadratic of the size of map-related variables. This computation complexity bottleneck explains why we maintain the map keyframes' poses instead of the map landmarks' positions in the state --- The size of the map landmarks is much larger than that of the map keyframes. However, even with map keyframe poses, the dimension of the augmented state can also be pretty high, which limits the real-time performance of the standard EKF.}


\textbf{Schmidt Update for map-based measurement} To further reduce the \textcolor{black}{computational} complexity, we employ the Schmidt-EKF to update the state partially. More specifically, we set $\mathbf{K}_{N_k} = \mathbf{0}$, leading to:
\textcolor{black}{
\begin{equation}
    \label{eq:schmidt_update_cov}
    \begin{aligned}
      \mathbf{P}_{k}&=\mathbf{P}_{k|k-1}-\\
      &\left[\begin{matrix} \mathbf{K}_{A_k}\mathbf{S}_k\mathbf{K}_{A_k}^{\top}& \mathbf{K}_{A_k}\textcolor{black}{\mathbf{H}_{k}^{*}} \left[\begin{matrix} \mathbf{P}_{AN_{k|k-1}}\\\mathbf{P}_{NN_{k|k-1}}
    \end{matrix}\right]\\
    \left[\begin{matrix} \mathbf{P}_{AN_{k|k-1}}\\\mathbf{P}_{NN_{k|k-1}}\\
    \end{matrix}\right]^{\top} \mathbf{H}_{k}^{*\top} \mathbf{K}_{A_k}^{\top}& \mathbf{0}\end{matrix}\right],
    \end{aligned}
\end{equation}}
and (\ref{eq:state_update}) will be divided into two parts:
\begin{equation}
\begin{aligned}
 \hat{\mathbf{x}}_{A_k}&=\hat{\mathbf{x}}_{A_{k|k-1}} + \mathbf{K}_{A_t}\mathbf{r}_{k}^{*},\\
    \label{eq:update_n}
    \hat{\mathbf{x}}_{N_k}&=\hat{\mathbf{x}}_{N_{k|k-1}}.
\end{aligned}
\end{equation}
It can be seen that the active part updating is identical to the standard EKF, while the nuisance part will not be updated.
Comparing (\ref{eq:expand_update_cov}) and (\ref{eq:schmidt_update_cov}), it is easy to see that:

\begin{equation}
\textcolor{black}{
    \mathbf{P}_{SKF}=\mathbf{P}_{EKF}+ \left[\begin{matrix}\mathbf{0}\\ \bar{\mathbf{K}}_{N_k}\end{matrix}\right] \mathbf{S}^{-1} \left[\begin{matrix}\mathbf{0}& \bar{\mathbf{K}}_{N_k}^{\top}\end{matrix}\right],
}
\end{equation}
where $\mathbf{P}_{SKF}$ and $\mathbf{P}_{EKF}$ are the updated covariance of the Schmidt-EKF and the standard EKF, respectively. As \textcolor{black}{$\mathbf{S}^{-1}_{k}$} is positive definite, \textcolor{black}{$\left[\begin{matrix}\mathbf{0}\\ \bar{\mathbf{K}}_{N_k}\end{matrix}\right] \mathbf{S}^{-1}_{k} \left[\begin{matrix}\mathbf{0}& \bar{\mathbf{K}}_{N_k}^{\top}\end{matrix}\right]$} must be positive semi-definite. Therefore, $\mathbf{P}_{SKF}-\mathbf{P}_{EKF}>\mathbf{0}$, indicating that the estimated state is consistent if the nuisance state estimation is consistent. As the consistency of the nuisance state is determined when it is removed from the state, the consistency of the state counts, which is discussed in the next section. Different from \cite{getout}, the cross-covariance of active state and map information is considered (cf. (\ref{eq:schmidt_update_cov})), with the cost of linear complexity $\mathcal{O}(m)$.

In summary, thanks to the keyframe-based map modeling, the map features null space projection, and the Schmidt updating, we can correct the drift of visual-inertial odometry by consistently using the map-based measurements while maintaining the system in \textcolor{black}{low computational complexity}.

\section{Observability Analysis and Improvement}

By considering the map uncertainty in the localization presented above, the covariance of the system can be estimated more reliably. Another source that may cause inconsistent estimation is linearization, which will break the original observability properties of the system and introduce spurious information along the unobservable directions of the state space. In this section, we analyze the observability properties of the augmented state system and demonstrate the improvement techniques for consistency and linearization error.

\subsection{Observability Analysis}
\label{sec:ob_analysis}
For simplicity, we assume the camera frame and the IMU frame are identical and consider one local feature. Besides, the IMU bias is neglected following \cite{gpsvio}. Therefore, we get the following simplified state vector:
\begin{equation}\label{eq:simple_state}
    \mathbf{x}_{s_k}=\begin{bmatrix}
    ^{I_k}\mathbf{q}_{L}^{\top}&^{L}\mathbf{v}_{I_k}^{\top}&^{L}\mathbf{p}_{I_k}^{\top}&^{L}\mathbf{p}_{f_k}^{\top}&^{G}\mathbf{q}_{L_k}^{\top}&^{G}\mathbf{p}_{L_k}
    \end{bmatrix}^{\top},
\end{equation}
\textcolor{black}{which includes the body pose $^{I_k}\mathbf{q}_{L}$, ${}^{L}\mathbf{p}_{I_k}$, velocity ${}^{L}\mathbf{v}_{I_k}$, a local feature $^{L}\mathbf{p}_{f_k}$ in the $L$ frame (original local inertial reference frame), and the relative transformation (augmented state) $^{G}\mathbf{q}_{L_k}$, $^{G}\mathbf{p}_{L_k}$ between the $L$ frame and $G$ frame (map or global reference frame) represented in $G$ frame. }

Denoting the state transition matrix of the augmented system from time step ${k-1}$ to $k$ as $\boldsymbol{\Phi}_{{k}|{k-1}}$, we have
\begin{equation}\label{ofvio}
\boldsymbol{\Phi}_{k|{0}} = \boldsymbol{\Phi}_{k|{k-1}}\ldots\boldsymbol{\Phi}_{1|0}.
\end{equation}
Accordingly, denoting the Jacobian matrix of the observation function of the visual-inertial odometry at time step $k$ as $\mathbf{H}_{L_k}$, and that of the map-based measurements as $\mathbf{H}_{G_k}$, we have the observability matrix following \cite{gpsvio} as
\begin{equation}\label{om}
\mathbf{M}\triangleq\begin{bmatrix}
\mathbf{H}_{L_0}\\
\mathbf{H}_{G_0}\\
\mathbf{H}_{L_1}\boldsymbol{\Phi}_{1|0}\\
\mathbf{H}_{G_1}\boldsymbol{\Phi}_{1|0}\\
\vdots\\
\mathbf{H}_{L_k}\boldsymbol{\Phi}_{k|0}\\
\mathbf{H}_{G_k}\boldsymbol{\Phi}_{k|0}\\
\end{bmatrix}
\triangleq
\begin{bmatrix}
\mathbf{M}_{L_0}\\
\mathbf{M}_{G_0}\\
\mathbf{M}_{L_1}\\
\mathbf{M}_{G_1}\\
\vdots\\
\mathbf{M}_{L_k}\\
\mathbf{M}_{G_k}\\
\end{bmatrix}.
\end{equation}

We first assume that the Jacobian matrices are evaluated at the groundtruth, which is ideal but can demonstrate the theoretic implication:

\textbf{Lemma 1}: (Ideal observability) The right null space $\mathcal{N}$ of the observability matrix $\mathbf{M}$, where the Jacobian matrices are evaluated ideally, is spanned by four directions as
\begin{equation}
\mathcal{N} = span\left(\begin{bmatrix}
^{I_0}{\mathbf{R}}_{L}\mathbf{g}&\mathbf{0}_{3}\\
(-^{L}{\mathbf{v}}_{I_0})_{\times}\mathbf{g}&\mathbf{0}_{3}\\
(-^{L}{\mathbf{p}}_{I_0})_{\times}\mathbf{g}&\mathbf{I}_{3}\\
(-^{L}{\mathbf{p}}_{f})_{\times}\mathbf{g}&\mathbf{I}_{3}\\
^{G}{\mathbf{R}}_{L}\mathbf{g}&\mathbf{0}_{3}\\
\mathbf{0}_{3\times1}&-^{G}\mathbf{R}_{L}\\
\end{bmatrix}\right).
\end{equation}

\textbf{Proof}: see Appendix \ref{sec:derive observability}.

However, in real practice, we cannot access the groundtruth poses, and can only evaluate the Jacobian matrices at some points with linearization error, which causes the spurious information and breaks the original observability of the system:

\textbf{Theorem 2}: (Real observability) The right null space $\mathcal{N}$ of the observability matrix $\mathbf{M}$, where the Jacobian matrices are evaluated at the changing estimated values, vanishes.

\textbf{Proof}: see Appendix \ref{sec:derive observability}.

\begin{figure}[t!]
    \centering
    \setlength{\abovecaptionskip}{0cm}
    \includegraphics[width=1\linewidth]{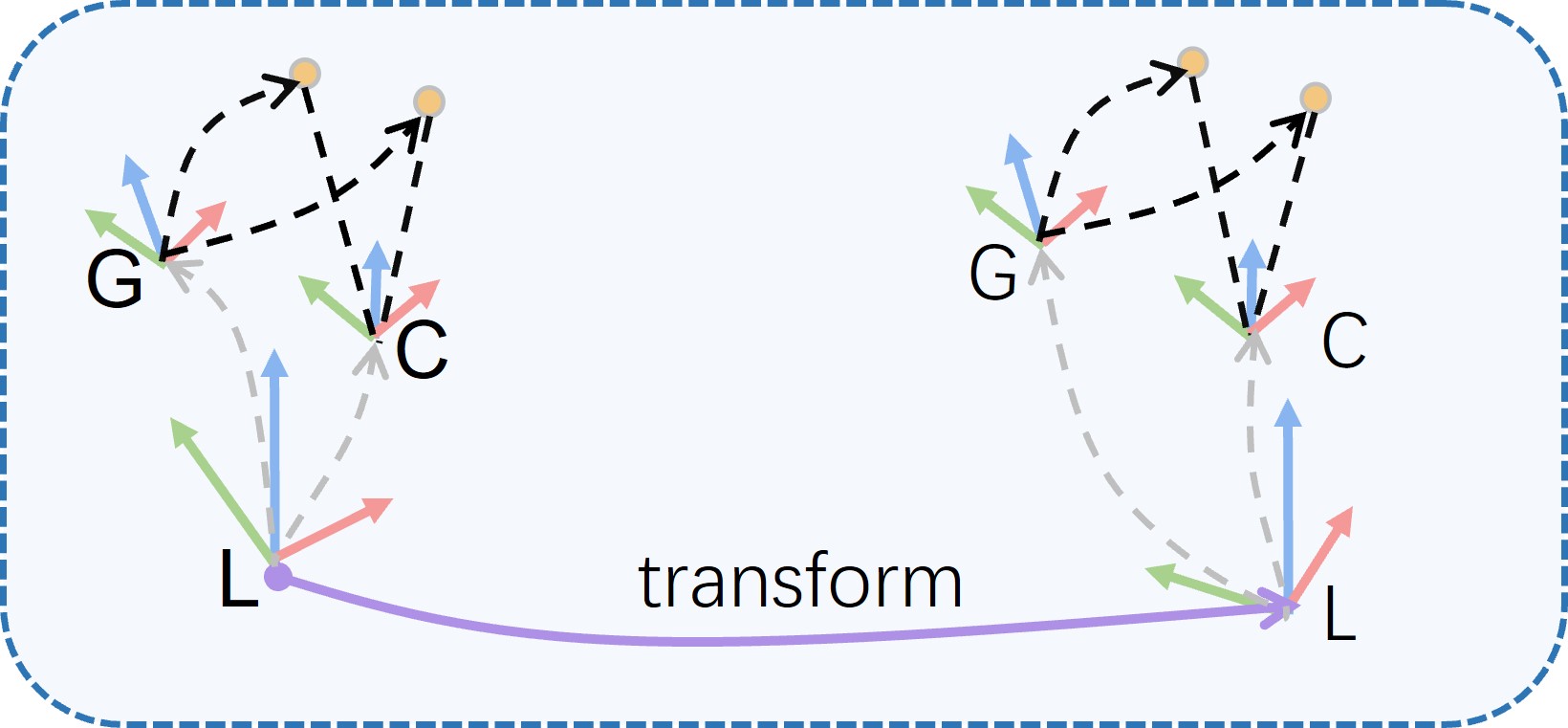}
     \caption{\textcolor{black}{Intuitive explanation towards observability. When a transformation is acted on the frame $L$, the relative transformations $^{G}\mathbf{T}_{L}$ and $^{L}\mathbf{T}_{C}$ will change, whereas the measurements towards the map landmarks (orange dots) remain unchanged.}}
     \label{fig:intuitive explanation}
     \vspace{-3mm}
\end{figure}

\textbf{Intuitive explanation} For the ideal system, the augmented system has the same unobservable dimensions as the standard visual-inertial system \cite{consistVIN}. This result is not surprising. We can understand the relation between the proposed result and previous works by regarding the augmented state as another $6$ degree of freedom (DoF) feature in the visual-inertial odometry frame and never being marginalized. With or without this $6$ DoF feature does not affect the unobservable subspace of the visual-inertial system.
Fig. \ref{fig:intuitive explanation} gives an intuitive explanation: The map landmarks (orange dots) in the frame $G$ are observed by the current image (frame $C$). The relative transformations $^{G}\mathbf{T}_{L}$ and $^{L}\mathbf{T}_{C}$ are represented as gray-dotted arrows. The map-based measurements are represented by the black-dotted lines, which will not change after applying a $4$ DoF (one for rotation along the gravity direction, three for translation) transformation. On the contrary, the estimated $^{G}\mathbf{T}_{L}$ and $^{L}\mathbf{T}_{C}$ can change with the applied $4$ DoF transformation. That is, $^{G}\mathbf{T}_{L}$ and $^{L}\mathbf{T}_{C}$ cannot be uniquely estimated using the map-based measurements. The applied arbitrary $4$ DoF transformation forms the so-called $4$ DoF unobservable subspace. However, different from the standard visual-inertial system in real scenarios, whose observability deficiency is along one direction (rotation around gravity direction)\cite{fej}, the observable direction along the translation of the augmented system also vanishes. \textcolor{black}{This is because the ever-changing linearization point of $^{G}\mathbf{R}_{L}$ introduces spurious information to the system and makes the system mistakenly thinks it can derive absolute values of the estimated state through the provided measurements. Obviously, the estimation is inconsistent\cite{converg and consist}}.


\textbf{First-Estimate Jacobian} From the analysis above, we aim to preserve observability to relieve inconsistency. We carefully extend the traditional first-estimate Jacobian (FEJ) \cite{fej} to the augmented system in three steps:
\begin{itemize}
	\item The propagation Jacobian matrix is evaluated at $\mathbf{x}_{k|k-1}$.
	\item The local feature observation Jacobian matrix is ever evaluated at first-estimate of $^{L}\mathbf{p}_{f}$ and $\mathbf{x}_{k|k-1}$.
	\item The map-based measurement Jacobian matrix is ever evaluated at first-estimate of $^{G}{\mathbf{p}}_{L}$, $^{G}\mathbf{R}_{L}$ and $\mathbf{x}_{k|k-1}$.
\end{itemize}
With these steps, the linearization points of $^{L}{\mathbf{p}}_{f}$ and $^{G}\mathbf{R}_{L}$ will never change. Therefore, the null space of the observability matrix $\mathbf{M}$ in real practice keeps the same as that in the ideal case. However, we emphasize that the employment of FEJ avoids changing the linearization point and improves the consistency, but the linearization error still exists.

\subsection{Observability Constrained Error Compensation}\label{relinear}

As shown in Sec. \ref{system}, in the update step, the observation function follows a general form $\mathbf{h}(\mathbf{x})$, and is linearizaed at $\hat{\mathbf{x}}$. We ignore the high order terms as
\begin{equation}
    \label{eq:linear_ob}
    \begin{aligned}
    \mathbf{z}&=\mathbf{h}(\mathbf{x})+\mathbf{n}\\
    &=\mathbf{h}(\hat{\mathbf{x}})+ \mathbf{H}_{\hat{\mathbf{x}}}(\mathbf{x}-\hat{\mathbf{x}})+o(\mathbf{x}^2)+\mathbf{n}\\
    &\approx \mathbf{h}(\hat{\mathbf{x}})+ \mathbf{H}_{\hat{\mathbf{x}}}(\mathbf{x}-\hat{\mathbf{x}})+\mathbf{n},
    \end{aligned}
\end{equation}
where $\mathbf{H}_{\hat{\mathbf{x}}}$ is the Jacobian matrix of the observation function linearized at $\hat{\mathbf{x}}$. This approximation is only accurate when $\hat{\mathbf{x}}$ is close to the true value. When the error between the estimated value and the true value is big, the poor approximation may lead to the failure of filtering.

\begin{figure}[t!]
    \centering
    \setlength{\abovecaptionskip}{0cm}
    \includegraphics[width=1\linewidth]{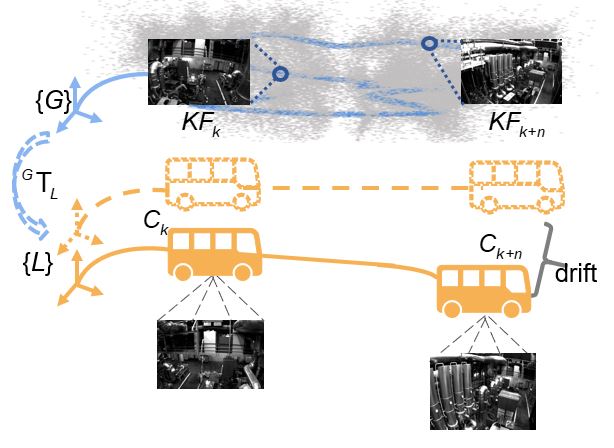}
     \caption{The illustration of the necessity of error compensated updating: the matched keyframes ($KF_k$,$KF_{k+n}$) from the map are represented in $\{G\}$; the orange dotted, and solid parts are the true and estimated state of the vehicle respectively, which are represented in $\{L\}$; the orange solid line parts are the estimated state of the vehicle. Due to the accumulation of the error, the error between the true value and the estimated value can be significant, which results in inaccurate linearization.}
     \label{fig:drift}
     \vspace{-5mm}
\end{figure}

\textbf{Absence of map-based measurements} In the observation function of map-based measurements (\ref{eq:global_observation_function}), the landmarks anchored in the matched keyframe are reprojected into the current image plane by a long concatenation of relative poses (cf. Fig. \ref{fig:coordinate}): The predicted observation of the landmark $^{{KF}_1}\mathbf{p}_{f_1}$  in the current image plane (black dot in Fig. \ref{fig:coordinate}) is derived from $^{G}\mathbf{T}_{{KF}_1}$, $^{G}\mathbf{T}_{L}$ and $^{L}\mathbf{T}_{{C}_k}$. Therefore, a large error in any of the related variables may cause the failure of filtering or at least introduce large errors to other variables in the update step due to the poor linearization approximation. 

Unfortunately, the map-based measurement can be absent for a long duration due to various reasons, say \textcolor{black}{the environment changes}\cite{jym}. Thus, the visual-inertial odometry drifts in this duration. When the map-based measurement is available again, the estimation of $^{L}\mathbf{T}_{{C}_k}$ has a large error, severely reducing the linearization accuracy of the observation function.

\textcolor{black}{Fig. \ref{fig:drift} provides a good insight of this problem. When there is a long-term absence of map-based measurement (from time step $k$ to time step $k+n$), due to the estimated error of $^{G}\mathbf{T}_{L}$ and the drift of the odometry, the reprojection errors of the map landmarks can be large, and the high order terms of the linearization cannot be ignored obviously.}   


\textbf{Limitation of re-linearzation} \textcolor{black}{Although improving the performance of the front-end (e.g., using more robust descriptors or using some techniques to reduce the feature false-matching rate \cite{SYBA2}) can mitigate the drift of the odometry, there still exists the problem that the linearization point is not accurate.} A simple idea to relieve the large error in the linearization point is to find a more accurate point. To be specific, when we get feature matchings between a map keyframe and the current image, we can calculate a relatively accurate pose ${}^{{KF}}\mathbf{T}_{{C}_k}$ using pose estimation technique. Based on this more accurate ${}^{{KF}}\mathbf{T}_{{C}_k}$, we can calculate a new $^{L}\mathbf{T}_{{C}_k}$, or a new $^{G}\mathbf{T}_{L}$ to replace the current one. This can be explained \textcolor{black}{as} we calculate the state through another chain of concatenation as shown in Fig. \ref{fig:coordinate}. The new value is then used to linearize the observation function (\ref{eq:global_observation_function}). However, as analyzed in Sec. \ref{sec:ob_analysis}, we employ FEJ to preserve the desired observability, which fixes the linearization point, preventing the observation function from being re-linearized.

\textbf{Large error compensation} Following FEJ constraint, when a map-based measurement is acquired, we have the observation function as
\begin{equation}\label{eq:poorlinear}
\mathbf{r} \triangleq \mathbf{z}-\mathbf{h}(\hat{\mathbf{x}}) \approx \mathbf{H}_{\mathbf{x}_{FEJ}}(\mathbf{x}-\hat{\mathbf{x}})+\mathbf{n}.
\end{equation}
where $\hat{\mathbf{x}}$ is the estimated value after the EKF propagation, $\mathbf{x}_{FEJ}$ is the first-estimate Jacobian matrix. We employ the re-calculated $^{G}\mathbf{T}_{L}$ above to replace the augmented state in $\hat{\mathbf{x}}$, denoted as $\mathbf{x}_r$. Then, we have 
\begin{equation}\label{eq:relinear1}
\mathbf{r}^{\prime} \triangleq \mathbf{z}-\mathbf{h}(\mathbf{x}_r) \approx \mathbf{H}_{\mathbf{x}_{FEJ}}(\mathbf{x}-\mathbf{x}_r)+\mathbf{n}.
\end{equation}
Note that we have the propagated value at $\hat{\mathbf{x}}$, which is different from the $\mathbf{x}_r$ in (\ref{eq:relinear1}). To address it, (\ref{eq:relinear1}) is rewritten as 
\begin{equation}\label{eq:comp}
\mathbf{r}^{\prime}  \approx \mathbf{H}_{\mathbf{x}_{FEJ}}(\mathbf{x}-\mathbf{x}_{r}+\hat{\mathbf{x}}-\hat{\mathbf{x}}) + \mathbf{n},
\end{equation}
and the compensated observation function is defined as:
\begin{equation}\label{eq:error_compensate}
^{4}\mathbf{r} \triangleq \mathbf{z}-\mathbf{h}(\mathbf{x}_r) - \mathbf{H}_{\mathbf{x}_{FEJ}}(\hat{\mathbf{x}} - \mathbf{x}_r) =\mathbf{H}_{\mathbf{x}_{FEJ}}\tilde{\mathbf{x}}+\mathbf{n}.
\end{equation}
\textcolor{black}{which is employed as the observation function for Schmidt update after the long-term absence of map-based measurements.} \textcolor{black}{The trigger of employing this compensated observation function is decided by a threshold, which is the average re-projection error of the map landmarks. In our experiments, this threshold is selected as $20$. The lower the threshold, the more frequently the error compensated updating mechanism is triggered, and vice versa.} In the experiments, we show that the new observation function (\ref{eq:error_compensate}) improves the accuracy of localization for large scenarios.

\section{Experimental Results}
\label{experiemt}
In this section, we would first demonstrate the setting and configuration of simulation, and datasets captured from an aerial vehicle, ground vehicles on campus, and urban areas (the first column of Fig. \ref{fig:dataset} shows the vehicles used to collect data). Then, extensive experimental results are given to validate the effectiveness of our proposed method. \textcolor{black}{All the experiments below are based on the state formulation (\ref{eq:full state}). The implementation of our proposed algorithm is based on the framework Open-VINS\cite{openvins}. Readers can refer to \cite{openvins} and its code for the detailed parameters. The settings of the newly introduced parameters are given in the \ref{sec:dataset}.}

\subsection{Datasets and Settings}
\label{sec:dataset}
\begin{figure}[t!]
\vspace{0.2cm}
    \centering
    \setlength{\abovecaptionskip}{0cm}
    \includegraphics[width=1\linewidth]{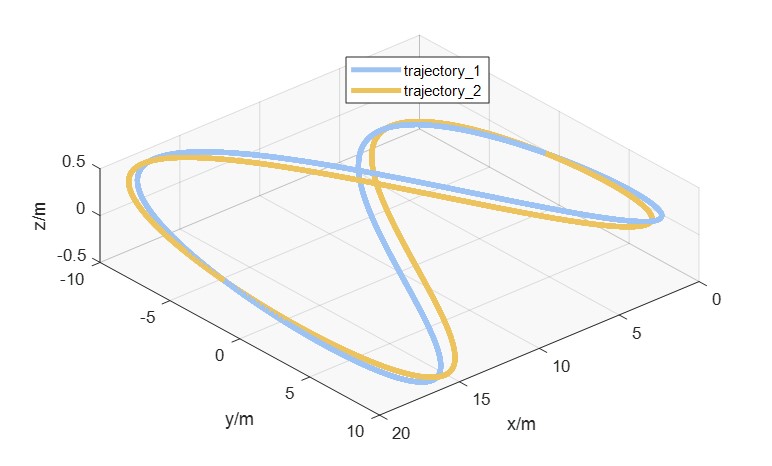}
     \caption{The trajectories used for simulation. The blue one is used to generate map information and the orange one is used to test algorithms.}
     \label{fig:eight_traj}
\end{figure}

\textbf{Simulation} We make some adaptations to the simulator of Open-VINS \cite{openvins} to evaluate the performance of the localization. We feed the simulator with a groundtruth trajectory shaped like ``number eight" which has a length of about 940m (cf. the orange trajectory of Fig.\ref{fig:eight_traj}). The map keyframe poses come from another trajectory, which has a similar shape as the previous trajectory (cf. the blue trajectory of Fig. \ref{fig:eight_traj} ). These two trajectories are denoted as EIGHT1 (the blue one) and EIGHT2 (the orange one), respectively.

To test the consistency of our proposed system, we artificially add the noise to the groundtruth of EIGHT1 to simulate the imperfect but real map keyframes. The position of the groundtruth is perturbed with the Gaussian white noise \textcolor{black}{$\mathbf{n}_p \sim \mathcal{N}(\mathbf{0},0.01\mathbf{I}_{3\times3} m^2)$}, and the orientation is perturbed with \textcolor{black}{$\mathbf{n}_o \sim \mathcal{N}(\mathbf{0},0.00025\mathbf{I}_{3\times3} rad^{2})$}. Therefore, the deviation for the position is $0.1 m$ and the deviation for the orientation is around $1 degree$. After the perturbation, the root-mean-squared error (RMSE) of the map trajectory of EIGHT1 is 0.179 $m$.

For the map matching information, we first randomly generate 3D landmarks, then we utilize the map-based observation function (\ref{eq:global_observation_function}) to reproject 3D landmarks into the perfect map keyframes and the frame in the running trajectory so that the 2D-2D matching features are obtained, after which we add Gaussian white noise to the 2D features. Finally, for each 3D landmark, we utilize the noisy 2D features and the noisy map keyframes to triangulate its 3D position as the estimated 3D landmark.

\begin{figure}[t]
\vspace{2mm}
    \centering

    \subfigure[EuRoC]{
    \begin{minipage}{1\linewidth}\centering
    \includegraphics[width=0.8\textwidth]{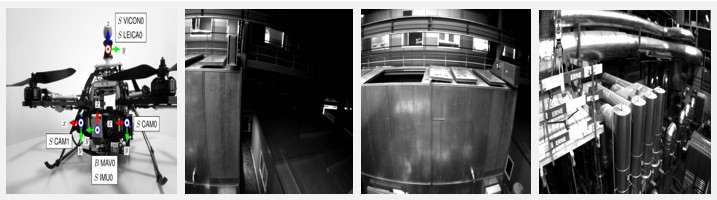}
    \end{minipage}
    }
    \subfigure[Kaist]{
    \begin{minipage}{1\linewidth}\centering
    \includegraphics[width=0.8\textwidth]{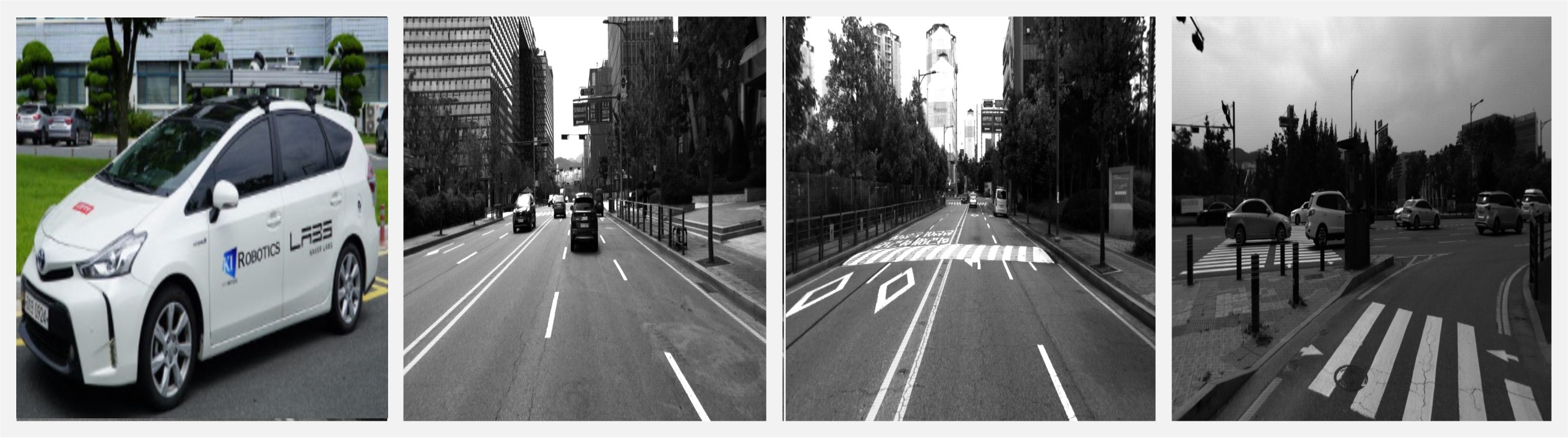}
    \end{minipage}
    }
    \subfigure[\textcolor{black}{4Seasons}]{
    \begin{minipage}{1\linewidth}\centering
    \includegraphics[width=0.8\textwidth]{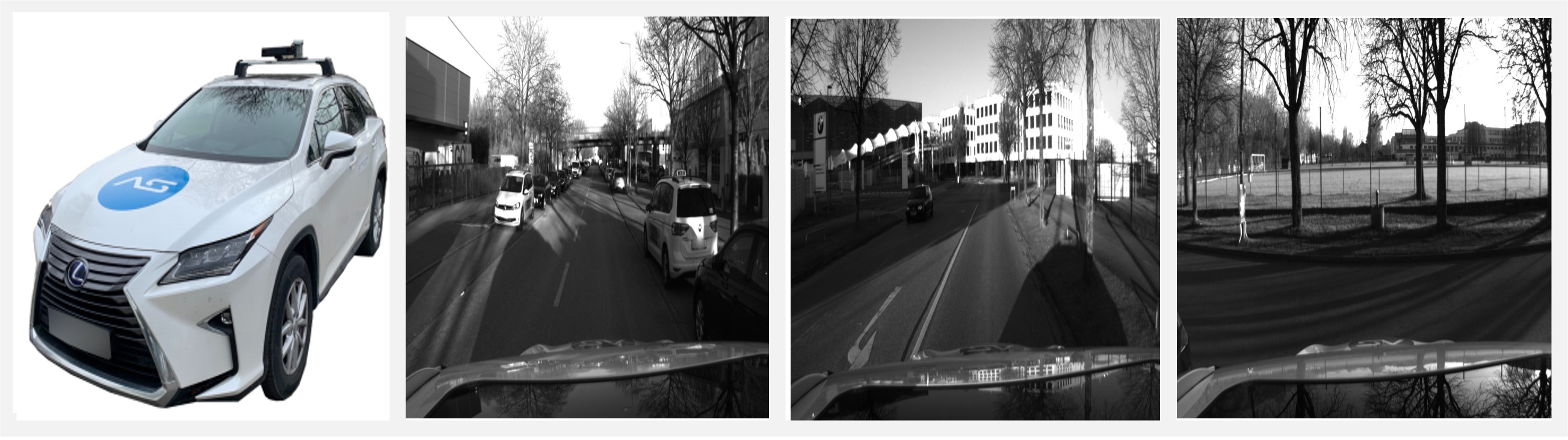}
    \end{minipage}
    }
    \subfigure[YQ]{
    \begin{minipage}{1\linewidth}\centering
    \includegraphics[width=0.8\textwidth]{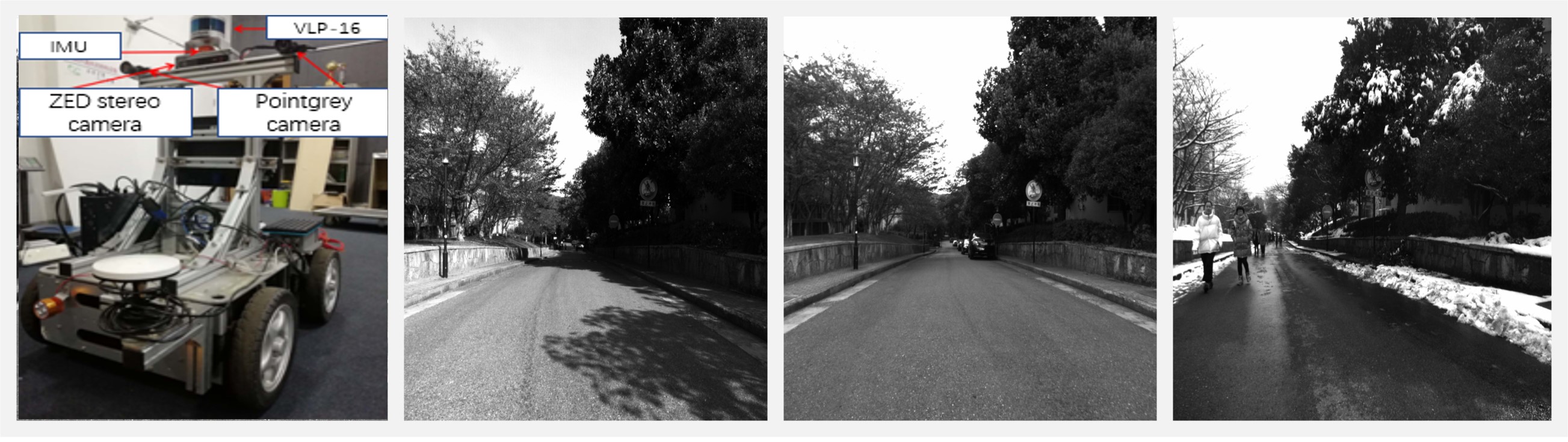}
    \end{minipage}
    }
    \setlength{\abovecaptionskip}{-0.3cm}
     \caption{\textcolor{black}{The images sampled from four datasets.}}
     \label{fig:dataset}
     \vspace{-3mm}
\end{figure}

\textbf{EuRoC} EuRoC\cite{euroc} is a visual-inertial dataset collected on-board a Micro Aerial Vehicle (MAV). This dataset contains scenarios of vicon room and machine hall. For our experiments, the sequences of machine hall (MH) are selected. The sequence MH01 is used to build the map, and the other sequences of MH (MH02-MH05) are used for localization. The map contains keyframes and 3D landmarks. As shown in dataset documents, the groundtruth poses have measurement errors. Therefore, we regard the groundtruth poses as the noisy map keyframes with the deviation of $1cm$ and $1^{\circ}$. The 3D landmarks are triangulated across multiple adjacent map images and the corresponding poses. The first row of Fig. \ref{fig:dataset} is some selected images from EuRoC.

\textbf{Kaist} Kaist\cite{kaist} is a dataset collected in the urban on-the-road environment with many moving vehicles (cf. the second row of Fig. \ref{fig:dataset}), so this is a challenging dataset causing drift in visual-inertial odometry. In the dataset, there are two sequences Urban38 and Urban39, having a large overlap. We employ Urban38 to build the map and Urban39 for localization testing. The length of the trajectory is 10.67km. The map keyframe and landmarks are calculated following that in EuRoC. As the groundtruth in this dataset is built upon outdoor RTK, we set the covariance of map keyframe pose as $0.1\mathbf{I}_{6\times6}$ \textcolor{black}{with the units of $m^2$ for the position part and $rad^2$ for the orientation part.} 

\textcolor{black}{\textbf{4Seasons} 4Seasons \cite{4seasons} is a dataset collected in different scenarios and under various weather conditions and illuminations. This dataset is for the research on VIO, global place recognition, and map-based re-localization, suitable to test our proposed algorithm. In this dataset, we select the first two sequences ($2020\text{-}03\text{-}24\_17\text{-}36\text{-}22$, denoted as Office-Loop-1, $2020\text{-}03\text{-}24\_17\text{-}45\text{-}31$, denoted as Office-Loop-2) from the collection called ``Office Loop", where the vehicle loops around an industrial area of the city. We pick these two sequences because the vehicle at the beginning of the sequences is static, which is necessary for the IMU initialization of our system. Besides, due to the high overlap of their trajectories and the similar illuminations, these two sequences are easier to detect matching images. We employ Office-Loop-1 to build the map and Office-Loop-2 for localization testing. The map keyframe and landmarks are generated following the procedure mentioned before. As the groundtruth in this dataset is built upon the RTK-GNSS that provide global positioning with up-to-centimeter accuracy \cite{4seasons}, we set the covariance of the map keyframe as $0.1\mathbf{I}_{6\times6}$ with the units of $m^2$ for the position part and $rad^2$ for the orientation part.}

\textbf{YQ} YQ\cite{dxq,YQ} is our own collected data in Yuquan Campus, Zhejiang University, China, by a four-wheel ground vehicle. This dataset aims at testing \textcolor{black}{our algorithm} in the off-the-road campus environment. It contains four sequences, YQ1-YQ4, where YQ1-YQ3 were recorded on three separate days with different weather in summer and YQ4 was collected in winter after snowing (cf. the fourth row of Fig. \ref{fig:dataset}). The changing weather and season severely degenerate the feature matching, causing the long absence of map-based measurements. For this dataset, we use YQ1 to build the map and YQ2-YQ4 to test the performance of the algorithms. The map setting is the same as that in Kaist.

\textbf{Feature matching} For all real-world datasets, the matching procedure is conducted as follows: We first utilize R2D2\cite{r2d2} to extract new features from the current query frame and match with features in map keyframes. The initial matching information based on the feature descriptor is then fed into a robust pose estimator in \cite{2entity} to generate accurate feature matchings.

\subsection{Ablation and Coefficients Analysis}
We first perform ablation studies to validate each proposed algorithm module substantially. We regard Open-VINS\cite{openvins} as the pure visual-inertial odometry baseline method. We call our basic module MVIL (map-based visual-inertial localization). As mentioned in Sec. \ref{global model},  we divide the map-based observation function as Single Matching (SM) and Multiple Matching (MM). To show the necessity of considering the map uncertainty, a setting that treats the map information as constant/perfect (MC) is used for comparison. Besides, whether using FEJ (Sec. \ref{sec:ob_analysis}) is also an essential factor for testing. To test the long absence of map-based measurements, the error compensated updating (ECU) mentioned in Sec. \ref{relinear} is also compared.

\textbf{Trajectory accuracy}  Simulation data are used to test the accuracy and consistency. The results of the ablation study are listed in Table \ref{tab:rmse_sim}. We can find that MVIL-MM-FEJ has the best result, which is in accordance with our theory: On the one hand, it has the correct null space of the observability matrix compared with MVIL w/o FEJ, and considers the uncertainty of the map compared with MVIL-MM-MC. On the other hand, compared with SM, multiple matching frames provide more constraints to improve the accuracy. Besides, when the map is treated as perfect (MVIL-MM-MC), the performance of the algorithm is even worse than that of pure odometry (Open-VINS), which demonstrates the necessity of taking the map uncertainty into account.

\begin{table}[t]
\caption{The RMSE/m of different methods on simulation data}
    \label{tab:rmse_sim}
    \centering
    \begin{tabular}{c|ccccc|c}
    \hline
    \makecell[c]{Open-VINS\cite{openvins}}&\makecell[c]{MVIL}&\makecell[c]{SM}&\makecell[c]{MM}&\makecell[c]{FEJ}&\makecell[c]{MC}&RMSE\\
    \hline
    \checkmark& & & & & &0.135\\
    &\checkmark&\checkmark& & & &0.131\\
    &\checkmark&\checkmark& &\checkmark& &0.113\\
    &\checkmark& &\checkmark& & &0.068\\
    &\checkmark& &\checkmark&\checkmark& &\textbf{0.057}\\
    &\checkmark& &\checkmark& &\checkmark&0.179\\
    \hline
    \end{tabular}
    \vspace{-3mm}
\end{table}

\begin{figure}[t]
\vspace{2mm}
    \centering

    \subfigure[trajectories on x-y plane]{
    \begin{minipage}{1\linewidth}\centering
    \includegraphics[width=1\textwidth]{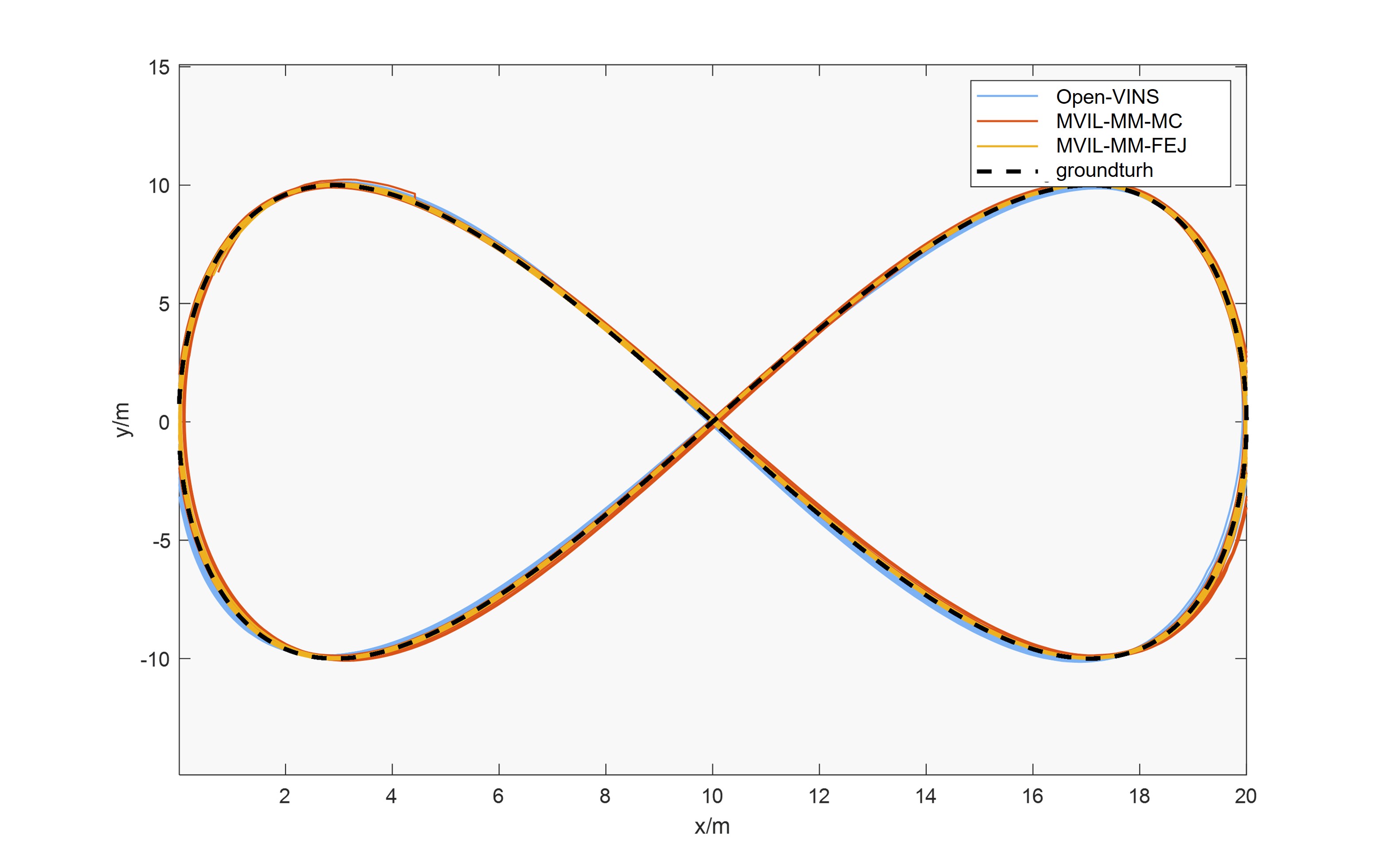}
    \end{minipage}
    }
    \subfigure[trajectories along z axis v.s. timestamps]{
    \begin{minipage}{1\linewidth}\centering
    \includegraphics[width=1\textwidth]{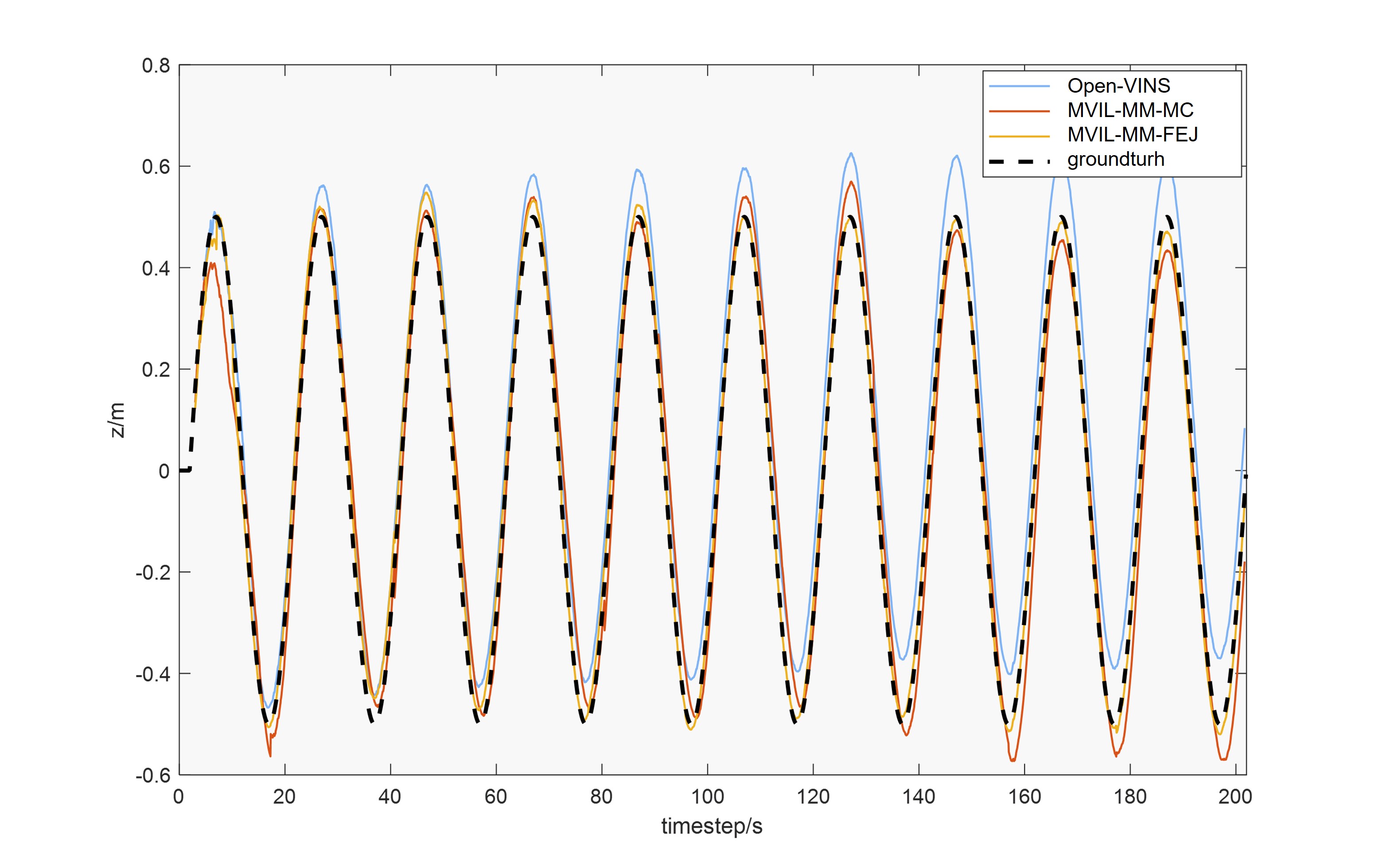}
    \end{minipage}
    }
     \caption{The trajectory comparison of different methods}
     \label{fig:eight_compare}
\end{figure}

Fig. \ref{fig:eight_compare} shows the trajectories derived from several configurations. The blue one is the trajectory from the Open-VINS odometry. We can see that the estimation drifts (especially along the $z$ axis), whereas our localization result (orange one for MVIL-MM-FEJ) is stably aligned with the groundtruth (black one). The red one is the result of the algorithm treating the map as perfect (MVIL-MM-MC). The reason for the worse performance is that the map information is not absolutely accurate. If we do not consider the uncertainty of the map, the estimator then overconfidently believes in the map information and incorrectly updates the state estimation.

\begin{figure}[htb]
\vspace{2mm}
\centering
\setlength{\abovecaptionskip}{0cm}

 \subfigure[IMU postion (${}^{L}\mathbf{p}_{I_k}$) error with 3-$\sigma$ bounds towards MVIL-MM-FEJ]{
    \begin{minipage}{0.9\linewidth}\centering
    \includegraphics[width=0.9\textwidth]{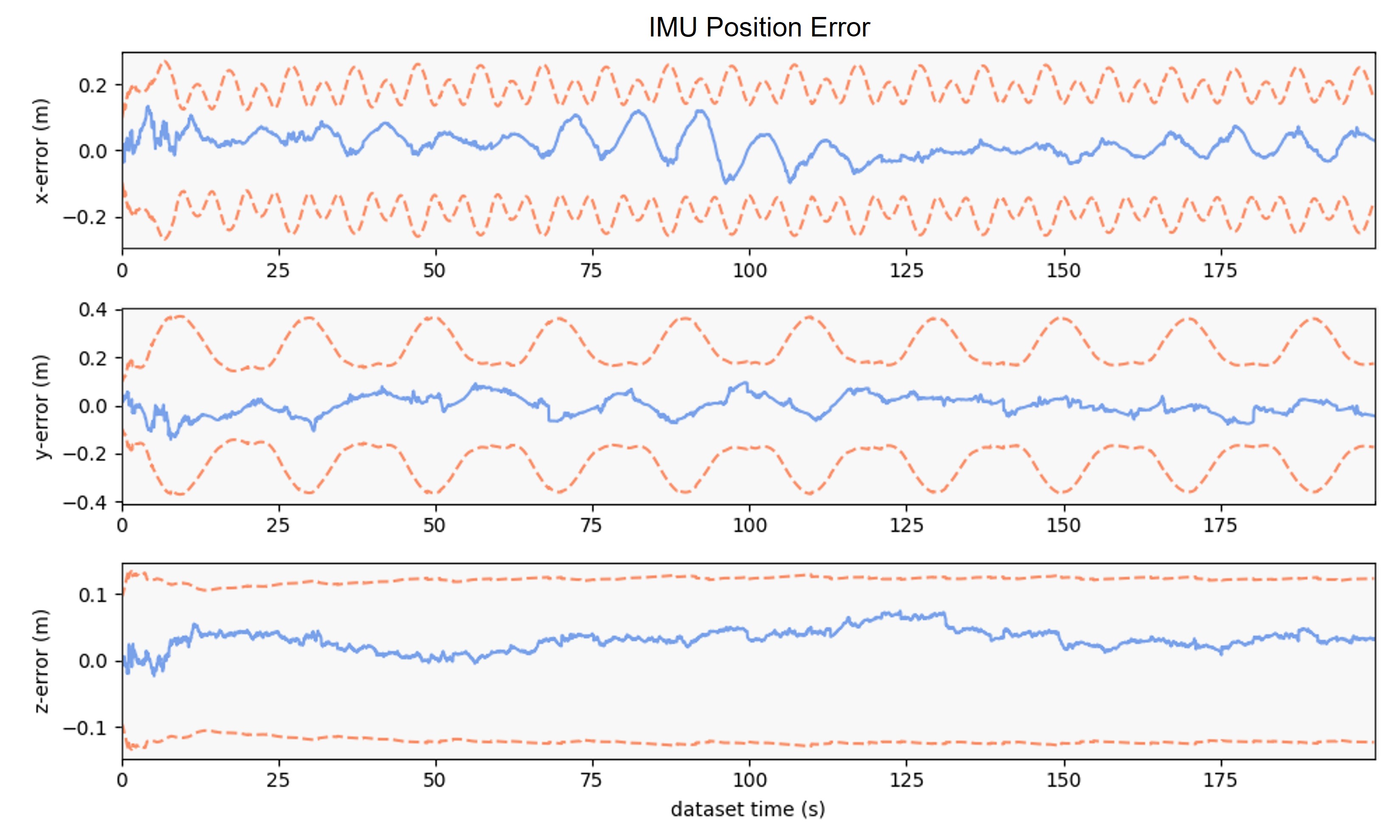}
    \end{minipage}
    }
\subfigure[Translation of Relative transform ($\mathbf{x}_t$) error with 3-$\sigma$ bounds towards MVIL-MM-FEJ]{
    \begin{minipage}{0.9\linewidth}\centering
    \includegraphics[width=0.9\textwidth]{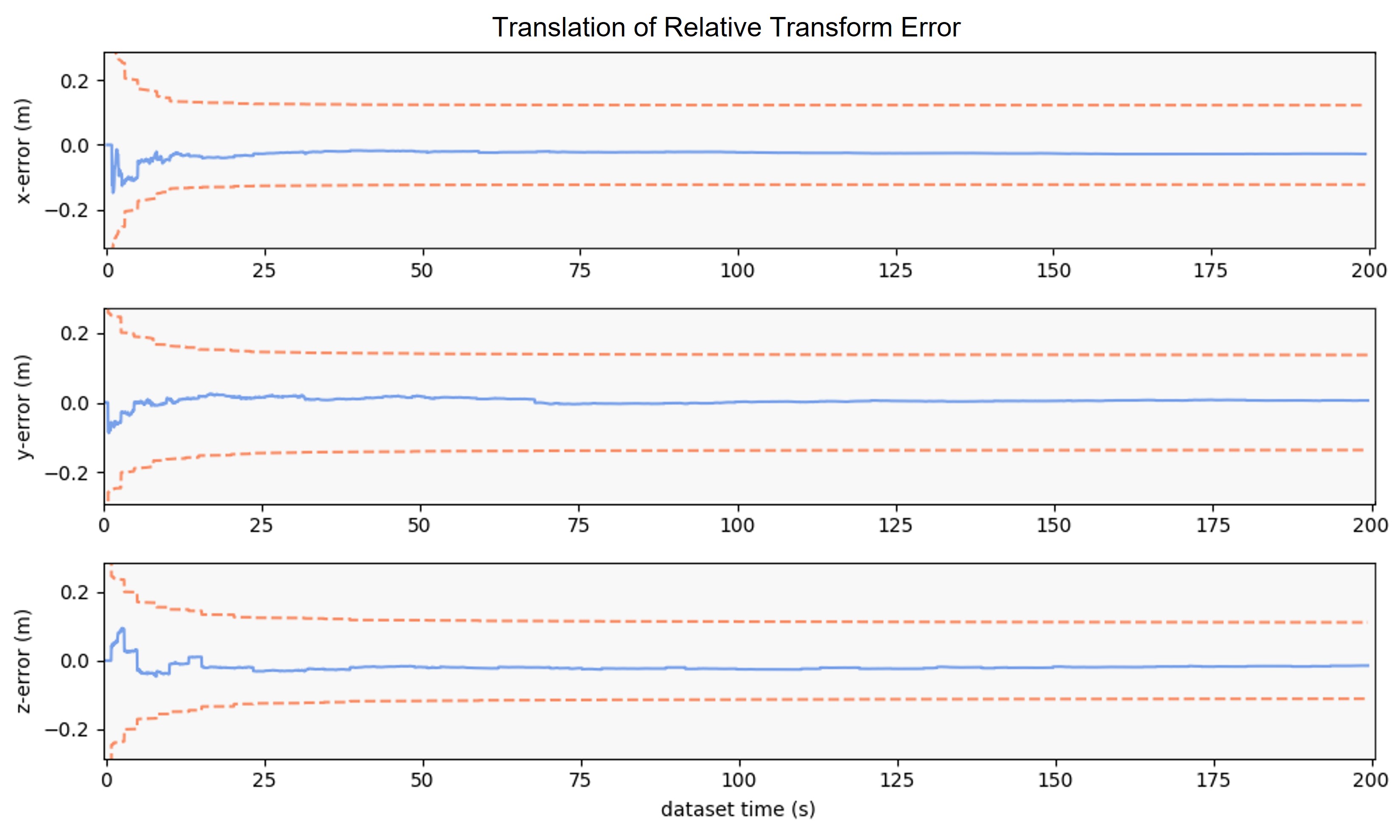}
    \end{minipage}
    }
\subfigure[IMU postion (${}^{L}\mathbf{p}_{I_k}$) error with 3-$\sigma$ bounds towards MVIL-MM-MC]{
    \begin{minipage}{0.9\linewidth}\centering
    \includegraphics[width=0.9\textwidth]{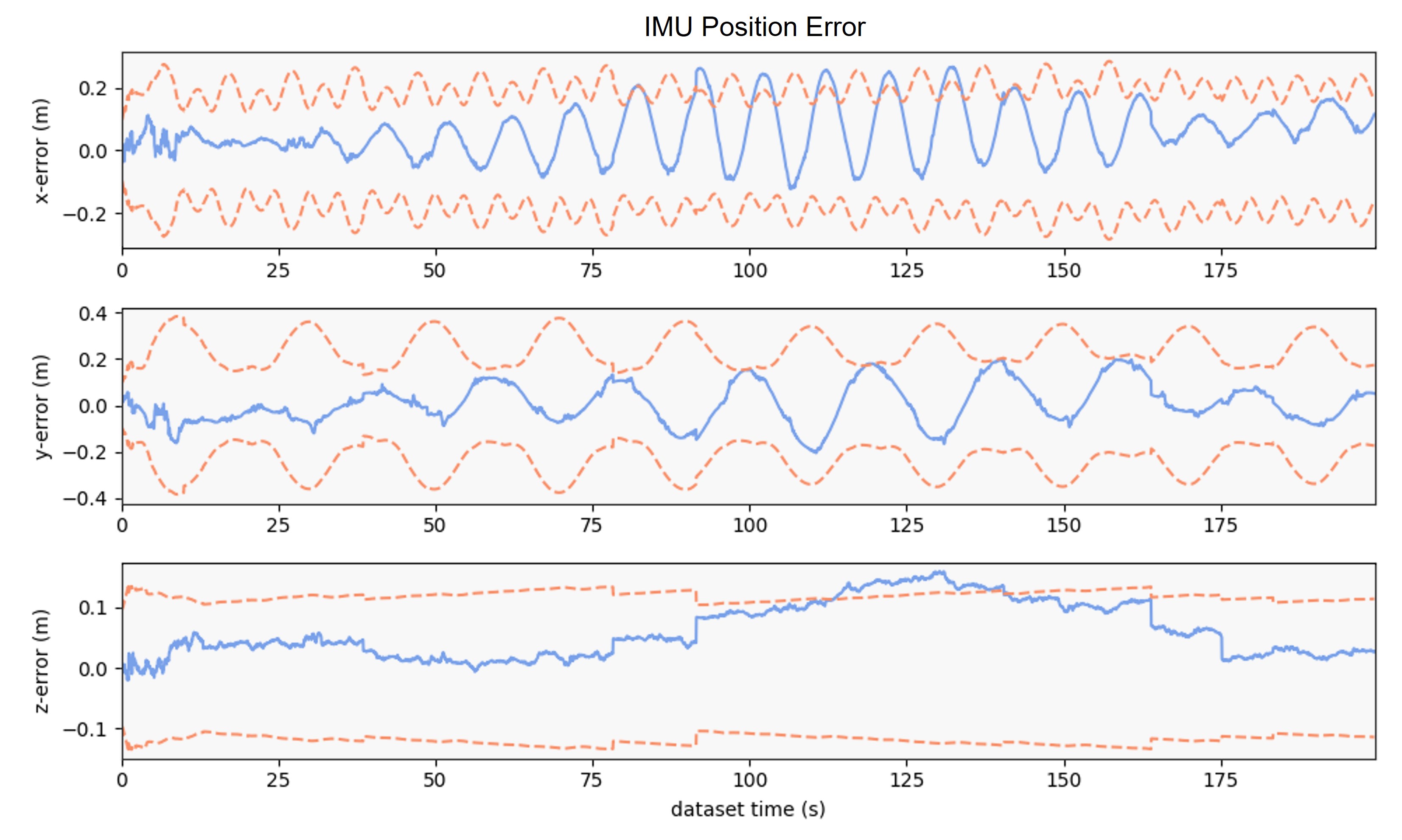}
    \end{minipage}
    }
\subfigure[Translation of Relative transform ($\mathbf{x}_t$) error with 3-$\sigma$ towards MVIL-MM-MC]{
    \begin{minipage}{0.9\linewidth}\centering
    \includegraphics[width=0.9\textwidth]{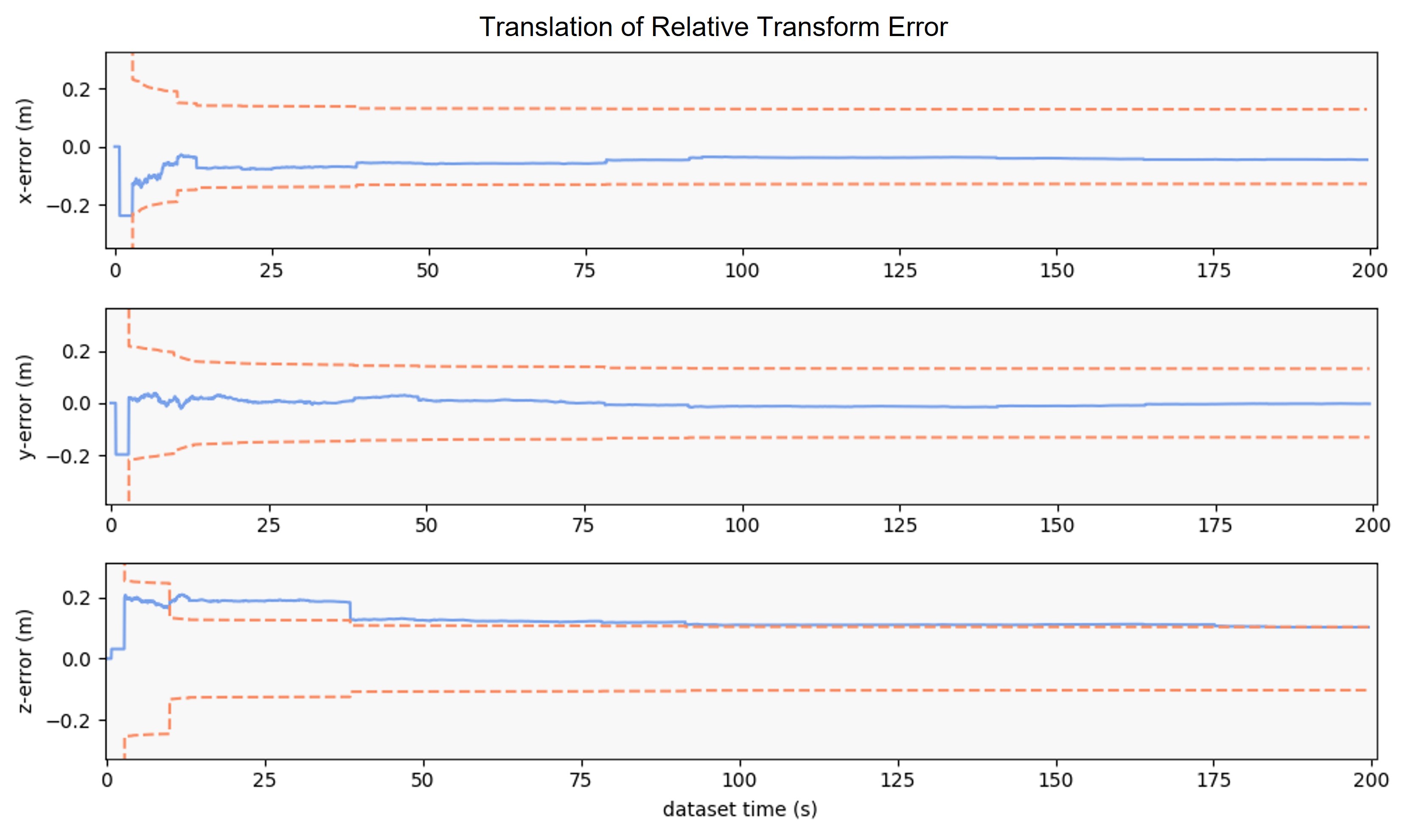}
    \end{minipage}
    }
\caption{The error and 3-$\sigma$ bounds towards MVIL-MM-FEJ/MC}
\label{fig:multi_3sigma}
\end{figure}

\textbf{Covariance reliability} To demonstrate the consistency of our proposed method, we follow the Monte Carlo evaluation in \cite{montecarlo}. We plot the evolution of estimation with 3-$\sigma$ bounds (cf. Fig. \ref{fig:multi_3sigma}). From the result, we conclude that our proposed algorithm has better consistency than the overconfident algorithms.

\begin{figure}[t]
    \centering
    \includegraphics[width=0.5\textwidth]{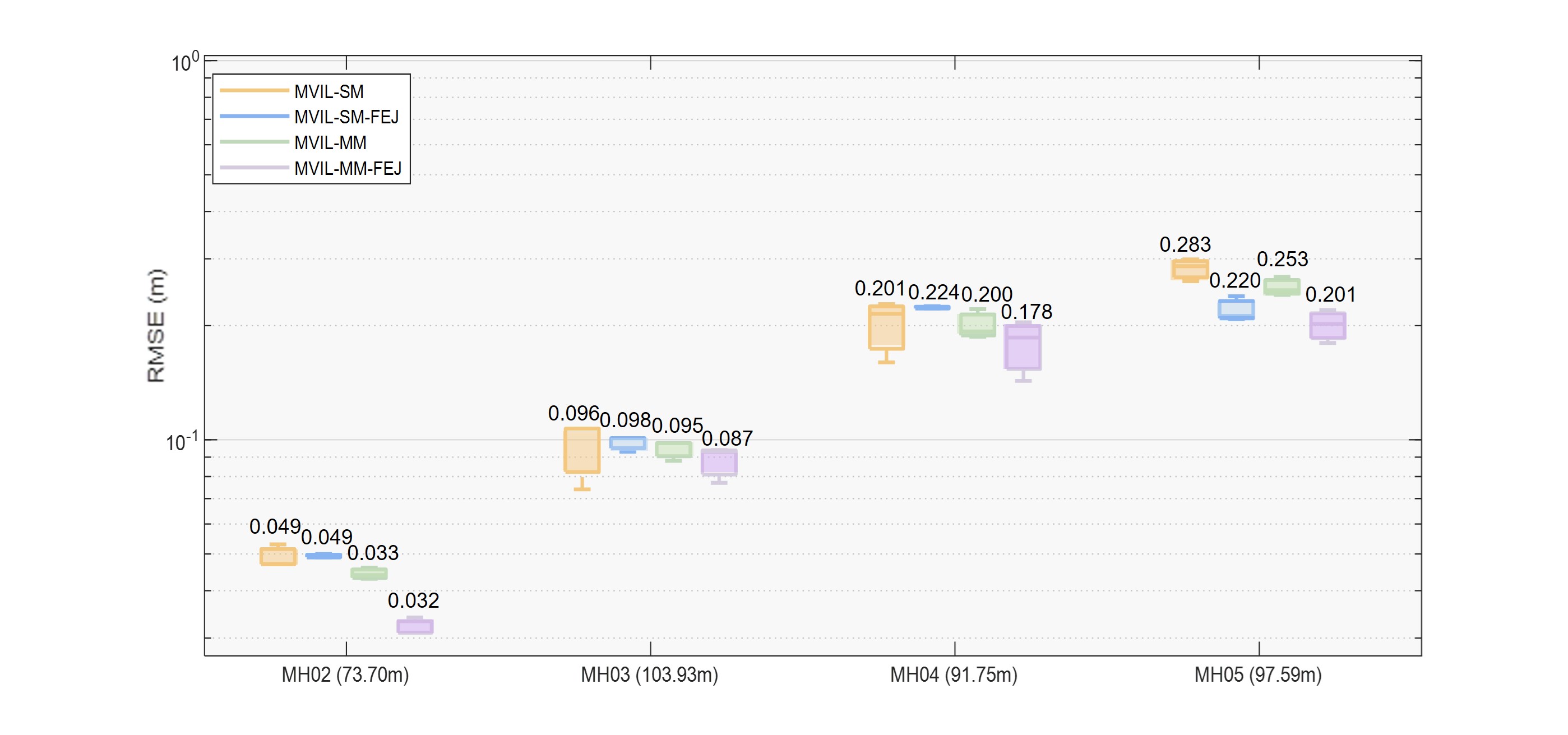}
     \caption{The RMSE/m of different methods on EuRoC\cite{euroc}}
     \label{fig:euroc_box}
\end{figure}

\textbf{Linearization strategy} We also compare our proposed methods with/without FEJ on EuRoC. The results are shown in Fig. \ref{fig:euroc_box}. The vertical axis is represented in the logarithm. We find that in most cases, methods that maintain the correct unobservable space have better results. Exceptions can be found for the SM setting. Since we use the first-estimate value to fix Jacobian linearization points, if this first-estimate value is inaccurate, the linearization error can act as the main error source, especially in such a short-term dataset. This phenomenon vanishes for the MM setting as we can use multiple matching information to calculate a better initial value of $^{G}\mathbf{T}_{L}$. 

\begin{figure}[t]
\vspace{2mm}
\centering
    \includegraphics[width=1\linewidth]{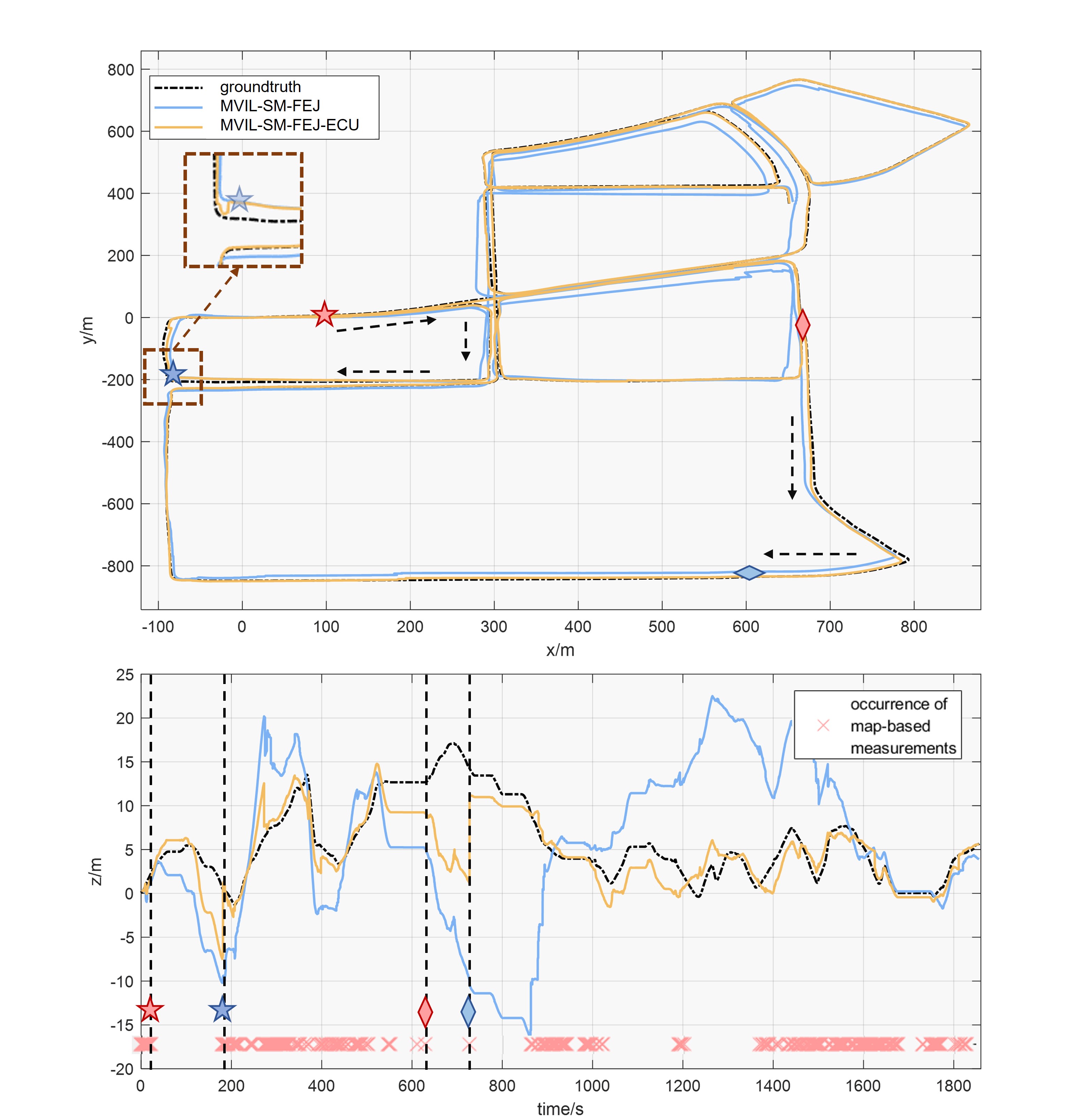}
     \caption{The trajectories derived from MVIL-SM-FEJ and MVIL-SM-FEJ-ECU.Top: trajectories on the $x-y$ plane; Bottom: trajectories along the $z$ axis v.s. time stamp; The details of the brown dotted square area is also given; The red and blue markers (stars and diamonds) represent the time when two adjacent matches occur (the matches at the red markers occur before the matches at the blue markers); The occurrence of map-based measurements is marked as red $\times$ at the bottom of the figure. }
     \label{fig:kaist_traj}
\end{figure}

\textbf{Error compensated updating} To show the advantage of our proposed error compensated updating (ECU) mechanism, we conduct experiments on the challenging Kaist dataset \cite{kaist}. The threshold for triggering ECU is set as 20 pixels. The comparison of different algorithms is given in Table \ref{tab:rmse_kaist}, and the trajectories derived from MVIL-SM-FEJ and MVIL-SM-FEJ-ECU are plotted in Fig. \ref{fig:kaist_traj}.
\begin{table}[thb]
    \caption{The RMSE/m of different methods on Kaist}
    \label{tab:rmse_kaist}
    \centering
    \begin{tabular}{c|ccccc|c}
    \hline
    IEKF\cite{iekf} & MVIL & SM & MM & FEJ & ECU & RMSE\\
    \hline
    &\checkmark&\checkmark& &\checkmark &  &17.965\\
    &\checkmark&\checkmark& &\checkmark &\checkmark&6.269\\
    \checkmark&\checkmark&\checkmark& &\checkmark & &8.574\\
    &\checkmark& &\checkmark&\checkmark& & 5.926\\
    &\checkmark& &\checkmark&\checkmark&\checkmark&\textbf{5.141}\\
    \hline
    \end{tabular}
\end{table}

From the results, we find that there is an obvious promotion in accuracy for both SM and MM scenarios when we employ the error compensated updating mechanism. We refer to Fig. \ref{fig:kaist_traj} for more illustration. In the figure, the pair of stars represents two contiguous matches (for example, the $i^{th}$ match in red and the ${i+1}^{th}$ match in blue), and so does the pair of diamonds. The motion direction of the car from the red mark to the blue mark is shown by black dotted arrows. The distance between the two stars or the two diamonds is far, which indicates the long absence of map-based measurements. Unfortunately, the odometry inevitably drifts in this interval, making \textcolor{black}{the first-order Taylor series approximation} of the observation function not good, as analyzed in Sec. \ref{relinear}. The effectiveness of ECU is highlighted in Fig. \ref{fig:kaist_traj}. We find that for the MVIL-SM-FEJ, the trajectory around the blue star (cf. the area of the brown dotted square) hardly changes when the map measurements are available again. On the contrary, the trajectory from MVIL-SM-FEJ-ECU shows an obvious correction. Besides, as shown in the bottom part of Fig. \ref{fig:kaist_traj}, the estimated values of $z$ also demonstrate a similar trend.

In addition, as the poor performance of MVIL-SM-FEJ is derived from the large linearization error, iterative EKF (IEKF) \cite{iekf} should be a useful way to improve the estimation. Therefore, in the experiment, we also compare our ECU with IEKF. The iteration number is assigned as $3$. Different from standard IEKF, to maintain the correct observability, the Jacobians are fixed with initial $^{L}\mathbf{T}_{C}$ and $^{G}\mathbf{T}_{L}$ as FEJ. From Table \ref{tab:rmse_kaist}, we find that our ECU performs better than IEKF. This can be explained by that ECU compensates for the error using the value from direct pose estimation, which better minimizes the observation function, so that the estimator can correct the estimated value significantly. On the other hand, IEKF compensates the error by iterating from the current imprecise estimated value, so that the estimator corrects poorly in the given time interval. In our implementation, ECU takes about half of the computation time of IEKF (around 3.7 ms vs. 6.9 ms). 

From Table \ref{tab:rmse_kaist}, algorithms with multiple matching frames also have good performance, which is consistent with our intuition. This could be explained by more map information provides more constraints on the current frame so that the significant drift could be corrected and bounded.

\textbf{Feature detection and description} In this part, we compare our proposed algorithms with different kinds of features under the setting of MVIL-FEJ-MM. The features used for comparison are R2D2\cite{r2d2}, \textcolor{black}{SYBA\cite{SYBA1}}, ORB\cite{orb}, SIFT\cite{sift} and SuperPoint\cite{sp}. \textcolor{black}{It should be noted that SYBA is an efficient feature descriptor that can be used in combination with various kinds of features\cite{SYBA1}. For our experiments, FAST\cite{fast} is used for feature extraction, and SYBA is used for feature description.} RMSE and the number of matched map keyframes (integer in brackets) are given in Table \ref{tab:rmse_euroc_feature}. The results show that the ORB feature matched the least number of map keyframes due to its lightweight descriptors, whereas the other \textcolor{black}{four} types of features matched a similar number of map keyframes. \textcolor{black}{It is worth noting that SYBA, as a traditional lightweight descriptor, has superior performance than ORB, or even better than SIFT.} \textcolor{black}{The learning-based features, R2D2 and SuperPoint, have more robust performance}.

\begin{table}
\caption{\textcolor{black}{The RMSE/m and the number of matched map keyframes with different features for MVIL-FEJ-MM on EuRoC}}
\label{tab:rmse_euroc_feature}
\setlength{\tabcolsep}{5pt}
\centering
\begin{tabular}{c|c|c|c|c}
\hline
Sequence&
\makecell[c]{MH02\\(73.70m)} &\makecell[c]{MH03\\(130.93m)} &\makecell[c]{MH04\\(91.75m)}& \makecell[c]{MH05\\(97.59m)} \\
\hline
\makecell[c]{R2D2 \cite{r2d2}} & \makecell[c]{\textbf{0.032} (205)} & \makecell[c]{0.087 (133)} & \makecell[c]{\textbf{0.178} (\textbf{63})} & \makecell[c]{0.201 (69)}  \\
\makecell[c]{SYBA \cite{SYBA1}}&\makecell[c]{0.034 (\textbf{319})}&\makecell[c]{\textbf{0.078}  (\textbf{169})}&\makecell[c]{0.306 (52)}&\makecell[c]{0.234 (38)}\\
\makecell[c]{ORB \cite{orb}}&\makecell[c]{0.065 (146)}&\makecell[c]{0.158 (84)}&\makecell[c]{0.324 (8)}&\makecell[c]{-}\\
\makecell[c]{SIFT \cite{sift}}&\makecell[c]{0.048 (207)}&\makecell[c]{0.084 (130)}&\makecell[c]{0.259 (58)} &\makecell[c]{0.251 (\textbf{71})}\\
\makecell[c]{SuperPoint\cite{sp}}&\makecell[c]{0.045 (205)}&\makecell[c]{0.100 (126)}&\makecell[c]{0.231 (55)}&\makecell[c]{\textbf{0.188} (58)}\\
\hline
\end{tabular}
\end{table}

\subsection{Benchmark with Comparative Methods}
In this part, we make a comparison between our proposed methods with the benchmark from Open-VINS\cite{openvins} and VINS-Fusion\cite{vinsmono,vinsgps} on EuRoC, Kaist, 4Seasons, and YQ. 


We regard Open-VINS as a pure odometry baseline to show the drift and validate the correction of map-based measurements. For VINS-Fusion, its localization mode is used. We keep its map \textcolor{black}{settings} the same as the ones in our method. The only difference is that VINS-Fusion ignores the uncertainty of the map. Therefore we do not set the covariance matrix of the map keyframes poses for VINS-Fusion. We record the localization result of VINS-Fusion by concatenating the odometry $^{L}\mathbf{T}_{k}$ and the estimated $^{G}\mathbf{T}_{L}$ instead of the optimized trajectory due to real-time causality.

\begin{table}
\caption{\textcolor{black}{The time gap (s) between consecutive map-based measurements for different datasets}}
\label{tab:map_frequency}
\setlength{\tabcolsep}{0.5pt}
\centering
\begin{tabular}{c|cccc|c|c|ccc}
\hline
Sequence&MH02&MH03&MH04&MH05&Urban39&\makecell[c]{Office-\\Loop-2}&YQ2&YQ3&YQ4\\
\hline
mean&0.274&0.486&1.107&1.063&4.317&0.868&1.580&0.986&1.479\\
min&0.250&0.250&0.250&0.250&0.997&0.264&0.500&0.497&0.502\\
max&2.500&27.250&46.500&55.250&168.602&44.979&114.984&90.069&123.558\\
std&0.185&2.172&6.294&6.669&16.684&2.446&7.090&3.474&8.331\\
\hline
\end{tabular}
\end{table}

\textcolor{black}{\textbf{Map-based measurement frequency} In Table \ref{tab:map_frequency}, we list the statistics of the time gap between consecutive map-based measurements (all data is measured in seconds), from which we can find that for most of the sequences, there is a long-term absence of the map-based measurement, which indicates that the odometry will have a large drift. This situation encourages us to utilize the ECU mechanism to improve the accuracy of the localization. The experimental results in Table \ref{tab:rmse_kaist} and Table \ref{tab:compare_all} also demonstrate the effectiveness of the ECU technique.
}

\textbf{Trajectory accuracy} The comparison of experimental results is shown in Table \ref{tab:compare_all}. \textcolor{black}{All the results are measured by RMSE and listed in the form of position (m) / orientation (degree).} The estimated trajectory of Open-VINS is aligned with the groundtruth by the first estimated pose. The other localization methods are compared without alignment. All the results are the average of three runs. Noting that as EuRoC is recorded in small scenes, the drift of the estimator is not distinct as that in the other three datasets, so the results of MVIL-SM/MM-EMU for EuRoC are omitted.

\begin{table*}[t]
\caption{\textcolor{black}{RMSE(m/degree) of different methods on different datasets}}
    \label{tab:compare_all}
    \setlength{\tabcolsep}{3pt}
    \centering
    \begin{tabular}{c|cccc|c|c|ccc}
    \hline
    Algorithms&\makecell[c]{MH02\\(73.70m)}&\makecell[c]{MH03\\(130.93m)}&\makecell[c]{MH04\\(91.75m)}&\makecell[c]{MH05\\(97.59m)}
    &\makecell[c]{Urban39\\(10.67km)}&\makecell[c]{Office-Loop-2\\(3.859km)}&\makecell[c]{YQ2\\(1.299km)}&\makecell[c]{YQ3\\(1.282km)}&\makecell[c]{YQ4\\(0.933km)}\\
    \hline
    Open-VINS\cite{openvins}&0.179/0.998&0.401/4.02&0.520/0.856&0.933/1.555&34.303/2.556&46.888/3.475&4.113/16.943&27.381/3.879&11.895/3.477\\
    VINS-Fusion\cite{vinsmono,vinsgps}&0.055/1.213&0.288/1.077&0.251/1.661&0.326/1.389&-&19.355/5.544&-&-&-\\
    MVIL-SM-FEJ&0.049/1.065&0.098/\textbf{1.192}&0.224/\textbf{0.795}&0.220/1.300&17.965/2.586&6.784/\textbf{2.696}&2.995/3.290&2.097/3.108&4.444/3.014\\
    MVIL-SM-FEJ-ECU&-&-&-&-&6.269/1.632&4.680/3.174   &1.724/\textbf{3.111}&1.751/\textbf{2.609}&3.570/\textbf{2.369}\\
    MVIL-MM-FEJ&\textbf{0.032/0.770}&\textbf{0.087}/1.398&\textbf{0.178}/0.816&\textbf{0.201/0.934}&5.926/1.708&6.355/3.207&2.029/3.245&1.429/3.214&3.314/2.427\\
    MVIL-MM-FEJ-ECU&-&-&-&-&\textbf{5.141/1.583}&\textbf{4.123}/2.951&\textbf{1.414}/3.175&\textbf{0.963}/3.076&\textbf{3.068}/2.498\\
    \hline
    \end{tabular}
\end{table*}

From Table \ref{tab:compare_all}, we can find that for EuRoC and 4Seasons, our proposed methods are more accurate than VINS-Fusion, because our framework considers the uncertainty of the map information and fuses the map information in a tightly-coupled manner. For large and challenging scenarios (Kaist and YQ), VINS-Fusion diverges due to the significant drifts and the overconfident belief in the noisy map. Even for the non-divergent situation (4Seasons), VINS-Fusion also gives poor performance. This is because VINS-Fusion neglects the uncertainty of the map information. Therefore when the map-based measurements are available, the estimator will highly rely on such measurements, which leads to poor performance. On the contrary, our proposed method can perform localization on all of these scenarios, and the error compensated updating mechanism improves the accuracy effectively as long term absence of map-based measurements occurs in both datasets \textcolor{black}{(map-based measurements can be absent for about three minutes/1km)}. \textcolor{black}{Combining multiple frames matching information (MM) and error compensated updating (ECU), our method outperforms the others in the position part and shows competitive performance in the orientation part.} 

\begin{figure}[h]
\vspace{2mm}
\centering
    \includegraphics[width=1\linewidth]{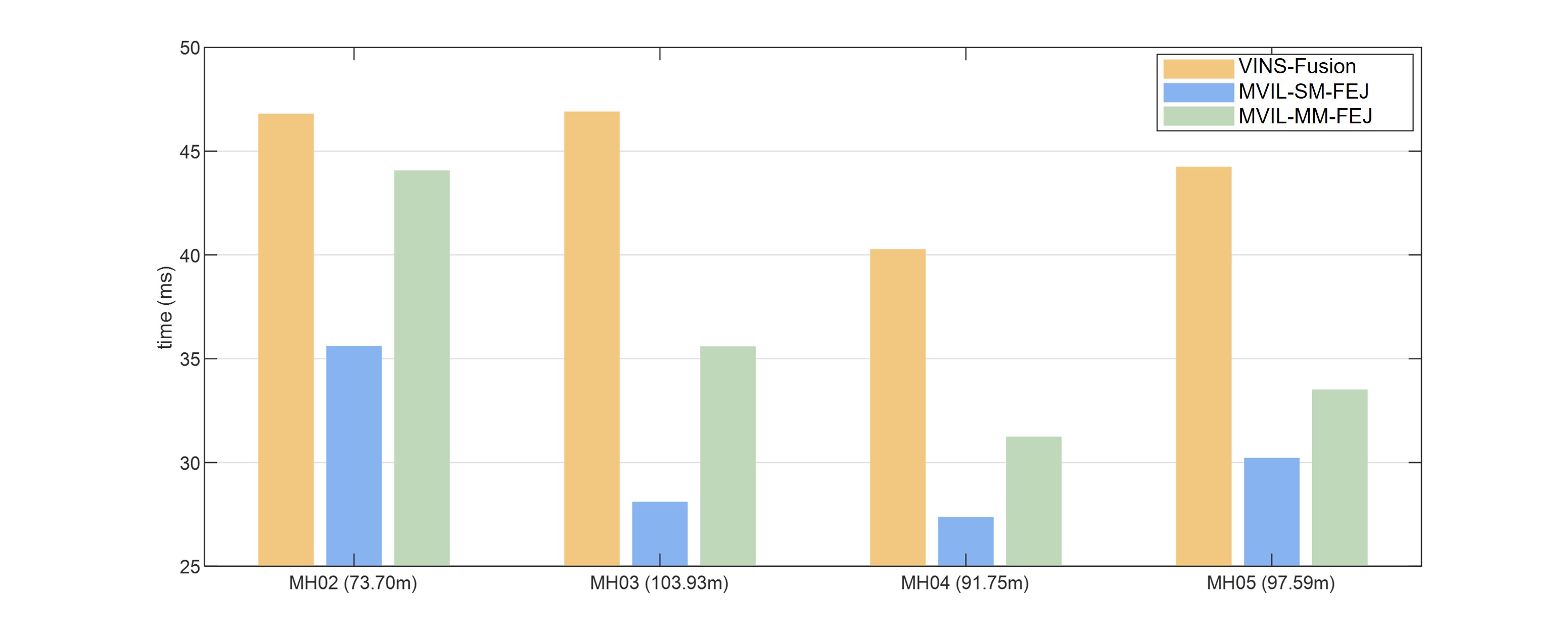}
     \caption{Time consumption (ms) of different methods on EuRoC\cite{euroc}}
     \label{fig:euroc_time}
\end{figure}

\textbf{Real-time efficiency} Efficiency is also essential for real-time localization. To show the efficiency of our proposed framework, the time consumption of VINS-Fusion and our method is shown in Fig. \ref{fig:euroc_time}. One can see that the filter-based solution is more time-saving than VINS-Fusion, thus more appropriate for on-board processing. 

\section{Conclusion}
\label{conclusion}
In this paper, we propose a filter-based localization system for autonomous vehicles, where map-based measurements are employed to bound the drift of \textcolor{black}{visual-inertial odometry.} The main advantages of the proposed solution are consistency and low \textcolor{black}{computational} complexity. To be specific, Schmidt-EKF is employed to reduce the \textcolor{black}{computational} complexity while maintaining the map uncertainty. Besides, theoretical analyses are shown on the observability of the augmented state formulation. Upon the analyses, \textcolor{black}{the first-estimate Jacobian technique is proposed to guarantee the dimension of the unobservable subspace being desired four by constraining the invariant linearization points. With the correct observability properties being maintained, the proposed localization algorithm can give consistent results.} Moreover, we introduce an observability constrained updating method to compensate for the significant drift after a long-term absence of map-based measurements, \textcolor{black}{which is useful especially for the long-term driving in large-scale scenes}. Finally, all the proposed modules are validated substantially in simulation and multi-platform real-world experiments.

\appendices
\section{Derivation of the unobservable space of the system}\label{sec:derive observability}
\textbf{Propagation matrix} With the state definition of (\ref{eq:simple_state}), the propagation functions are given as:
\begin{equation}
    \left\{\begin{aligned} \textcolor{black}{^{I_k}\dot{\mathbf{q}}_{L}}&=\frac{1}{2}\Omega(\boldsymbol{\omega}_{m_k}-\mathbf{b}_{g_k}-\mathbf{n}_{g_k})\textcolor{black}{{}^{I_k}\mathbf{q}_{L}}&\\
    ^{L}\dot{\mathbf{v}}_{I_k}&=C(\textcolor{black}{^{I_k}{\mathbf{q}}_{L}})^{\top}(\mathbf{a}_{m_k}-\mathbf{b}_{a_k}-\mathbf{n}_{a_k})+\mathbf{g}\\
    ^{L}\dot{\mathbf{p}}_{I_k}&=\textcolor{black}{{}^{L}\mathbf{v}_{I_k}} \;\;\;\;
    ^{L}\dot{\mathbf{p}}_{f_k}=\mathbf{0}\\
    ^{G_k}\dot{\mathbf{q}}_{L}&=\mathbf{0}\;\;\;\;\;\;\;\;\,
    ^{G_k}\dot{\mathbf{p}}_{L}=\mathbf{0}\\
    \end{aligned}
    \right. ,
\end{equation}
where $C(\cdot)$ transforms a quaternion to a rotation matrix. \textcolor{black}{$\boldsymbol{\omega}_{m_k}$ and $\mathbf{a}_{m_k}$ are the measurements from IMU in frame $L$, $\mathbf{b}_{g_k}$ and $\mathbf{b}_{a_k}$ are the bias in frame $L$, $\mathbf{n}_{g_k}$ and $\mathbf{n}_{a_k}$ are the noises (cf. (\ref{eq:imu_state}) and (\ref{eq:state_prop})). $\mathbf{g}=\begin{bmatrix}0&0&-9.81\end{bmatrix}^{\top}m/s^2$ is the gravitational acceleration in the $L$ frame.} \textcolor{black}{For a vector $\boldsymbol{\omega}\in \mathbb{R}^{3}$, we have \begin{equation} \nonumber
    \Omega(\boldsymbol{\omega})=\begin{bmatrix}
    -\boldsymbol{\omega}_{\times}&\boldsymbol{\omega}\\
    \boldsymbol{\omega}^{\top}&0
    \end{bmatrix}.
\end{equation}}

Following the derivation of \cite{gpsvio}, we could get that the state transition matrix is:
\begin{equation} \label{eq:transition matrix}
    \boldsymbol{\Phi}_{{k+1}|{k}}=
    \begin{bmatrix}
    \boldsymbol{\Phi}_1&\mathbf{0}&\mathbf{0}&\mathbf{0}&\mathbf{0}&\mathbf{0}\\
   \boldsymbol{\Phi}_2&\mathbf{I}&\mathbf{0}&\mathbf{0}&\mathbf{0}&\mathbf{0}\\
    \boldsymbol{\Phi}_3&\mathbf{I}\Delta &\mathbf{I}&\mathbf{0}&\mathbf{0}&\mathbf{0}\\
    \mathbf{0}&\mathbf{0}&\mathbf{0}&\mathbf{I}&\mathbf{0}&\mathbf{0}\\
     \mathbf{0}&\mathbf{0}&\mathbf{0}&\mathbf{0}&\mathbf{I}&\mathbf{0}\\
      \mathbf{0}&\mathbf{0}&\mathbf{0}&\mathbf{0}&\mathbf{0}&\mathbf{I}\\
    \end{bmatrix},
\end{equation}
where
\begin{equation}
    \nonumber
    \begin{aligned}
      \boldsymbol{\Phi}_1&={}^{I_{k+1}}\hat{\mathbf{R}}_{L}{}^{I_{k}}\hat{\mathbf{R}}_{L}^{\top},\\
      \boldsymbol{\Phi}_2&=-(^{L}\hat{\mathbf{v}}_{I_{k+1}}-^{L}\hat{\mathbf{v}}_{I_{k}}+\mathbf{g}\Delta )_{\times}{}^{I_k}\hat{\mathbf{R}}_{L}^{\top},\\
      \boldsymbol{\Phi}_3&=-(^{L}\hat{\mathbf{p}}_{I_{k+1}}-^{L}\hat{\mathbf{p}}_{I_{k}}-^{L}\hat{\mathbf{v}}_{I_{k}}\Delta+\frac{1}{2}\mathbf{g}\Delta ^{2})_{\times}{}^{I_k}\hat{\mathbf{R}}_{L}^{\top}.
    \end{aligned}
\end{equation}
$\Delta$ is one time step from time step $k$ to time step $k+1$. The operation $(\cdot)_{\times}$ transforms a vector into a skew symmetric matrix.

\vspace{0.3cm}
\textbf{Jacobian matrix of local feature measurements} Suppose the local feature is $^{L}{\mathbf{p}}_{f}$, then we have the following observation model:
\begin{equation}
    \textcolor{black}{{}^{L}\mathbf{z}_{k}}=h(^{I_k}\mathbf{R}_{L}(^{L}\mathbf{p}_{f_k}-^{L}\mathbf{p}_{I_k}))+\textcolor{black}{{}^{L}\mathbf{n}_{k}},
\end{equation}
where $h(\cdot)$ is the projection function, \textcolor{black}{${}^{L}\mathbf{z}_{k}$ and ${}^{L}\mathbf{n}_{k}$ are the local feature measurement and the zero-mean Gaussian measurement noise, respectively}. Here, for simplicity, the extrinsics between camera and IMU is set Identity.
The Jacobians of the above function are given by:
\begin{equation} \label{eq:HLk}
    \mathbf{H}_{L_k}=\mathbf{H}_{\pi_k}\begin{bmatrix}
    \mathbf{H}_{L1}&\mathbf{0}&-^{I_k}\hat{\mathbf{R}}_{L}&^{I_k}\hat{\mathbf{R}}_{L}&\mathbf{0}&\mathbf{0}
    \end{bmatrix},
\end{equation}
where $\mathbf{H}_{\pi_k}$ is the Jacobians of the projection function and
\begin{equation}\nonumber
\mathbf{H}_{L1} = (^{I_k}\hat{\mathbf{R}}_{L}(^{L}\hat{\mathbf{p}}_{f_k}-^{L}\hat{\mathbf{p}}_{I_k}))_{\times}.    
\end{equation}

\vspace{0.3cm}
\textbf{Jacobian matrix of map-based measurements}
Suppose there is a given feature with the position in the map reference system $^{G}\mathbf{p}_{F}$, then the observation function related to this feature is given by:
\begin{equation}
    \textcolor{black}{^{G}\mathbf{z}_{k}}=h\left[^{I_k}\mathbf{R}_{L}(^{G_k}\mathbf{R}_{L}^{\top}(^{G}\mathbf{p}_{F}-{}^{G_k}\mathbf{p}_{L})-{}^{L}\mathbf{p}_{I_k})\right]+\textcolor{black}{{}^{G}\mathbf{n}_{k}},
\end{equation}
\textcolor{black}{where $^{G}\mathbf{z}_{k}$ and ${}^{G}\mathbf{n}_{k}$ are the map-based measurement and the measurement noise, respectively.}
The Jacobians of the above function are given by:
\begin{equation}\label{eq:HGk}
    \mathbf{H}_{G_k}=\mathbf{H}_{\pi_k}^{'}\begin{bmatrix}
    \mathbf{H}_{G1}&\mathbf{0}&-^{I_k}\hat{\mathbf{R}}_{L}&\mathbf{0}&\mathbf{H}_{G2}&-^{I_k}\hat{\mathbf{R}}_{L}{}^{G_k}\hat{\mathbf{R}}_{L}^{\top}\\
    \end{bmatrix},
\end{equation}
where $\mathbf{H}_{\pi_k}$ is the Jacobians of the projection function. The superscript $'$ is just for distinguishing from local observation. 
\begin{equation}
    \nonumber
    \begin{aligned}
      \mathbf{H}_{G1}&=\left[^{I_k}\hat{\mathbf{R}}_{L}(^{G_k}\hat{\mathbf{R}}_{L}^{\top}(^{G}\mathbf{p}_{F}-{}^{G_k}\hat{\mathbf{p}}_{L})-{}^{L}\hat{\mathbf{p}}_{I_k})\right]_{\times},\\
      \mathbf{H}_{G2}&=- {}^{I_k}\hat{\mathbf{R}}_{L}{}^{G}\hat{\mathbf{R}}_{L}^{\top}(^{G}\mathbf{p}_{F}-{}^{G_k}\hat{\mathbf{p}}_{L})_{\times}.
    \end{aligned}
\end{equation}

\vspace{0.3cm}
\textbf{Observability matrix}
Assume at time step $k-1$, we have a state denoted as $\mathbf{x}_{k-1|k-1}$. After dynamics propagation, we get the state $\mathbf{x}_{k|k-1}$. After update step, we get the state at time step $k$, i.e. $\mathbf{x}_{k|k}$.

The following derivations fall into two cases: (a) one is that the prediction state $\mathbf{x}_{k|k-1}$ is equal to update state $\mathbf{x}_{k|k}$, which is ideal situation. In this situation, we denote both $\mathbf{x}_{k|k-1}$ and $\mathbf{x}_{k|k}$ as $\mathbf{x}_{k}$. (b) The other one is that the prediction state $\mathbf{x}_{k|k-1}$ is not equal to update state $\mathbf{x}_{k|k}$.

$\bullet$ For case (a):
\begin{equation} \label{eq:prop_matrix}
\begin{aligned}
      \boldsymbol{\Phi}_{k|0}&=\boldsymbol{\Phi}_{k|{k-1}}\dots \boldsymbol{\Phi}_{1|0}\\
      &= \begin{bmatrix}
   \boldsymbol{\Phi}(1)&\mathbf{0}&\mathbf{0}&\mathbf{0}&\mathbf{0}&\mathbf{0}\\
  \boldsymbol{\Phi}(2) &\mathbf{I}&\mathbf{0}&\mathbf{0}&\mathbf{0}&\mathbf{0}\\
   \boldsymbol{\Phi}(3) &\mathbf{I}\Delta_k &\mathbf{I}&\mathbf{0}&\mathbf{0}&\mathbf{0}\\
    \mathbf{0}&\mathbf{0}&\mathbf{0}&\mathbf{I}&\mathbf{0}&\mathbf{0}\\
     \mathbf{0}&\mathbf{0}&\mathbf{0}&\mathbf{0}&\mathbf{I}&\mathbf{0}\\
      \mathbf{0}&\mathbf{0}&\mathbf{0}&\mathbf{0}&\mathbf{0}&\mathbf{I}\\
    \end{bmatrix},\\
\end{aligned}
\end{equation}
where $\Delta_k$ is $k$ time steps from time setp $0$ to time setp $k$,
\begin{equation}
    \nonumber
    \begin{aligned}
      \boldsymbol{\Phi}(1)&={}^{I_{k}}\hat{\mathbf{R}}_{L}{}^{I_{0}}\hat{\mathbf{R}}_{L}^{\top},\\
      \boldsymbol{\Phi}(2)&=- (^{L}\hat{\mathbf{v}}_{I_{k}}-^{L}\hat{\mathbf{v}}_{I_{0}}+\mathbf{g}\Delta_ k)_{\times}{}^{I_0}\hat{\mathbf{R}}_{L}^{\top},\\
      \boldsymbol{\Phi}(3)&=-(^{L}\hat{\mathbf{p}}_{I_{k}}-^{L}\hat{\mathbf{p}}_{I_{0}}-^{L}\hat{\mathbf{v}}_{I_{0}}\Delta _k+\frac{1}{2}\mathbf{g}\Delta _k^{2})_{\times}{}^{I_0}\hat{\mathbf{R}}_{L}^{\top}.
    \end{aligned}
\end{equation}

Recalling the Jacobians of the local-feature-based and map-based observation function, we could get the observability matrix
\begin{equation}
\begin{aligned}
       \mathbf{M}&=\begin{bmatrix}
    \mathbf{H}_{L_0}\\
    \mathbf{H}_{G_0}\\
    \mathbf{H}_{L_1}\boldsymbol{\Phi}_{1|0}\\
    \mathbf{H}_{G_1}\boldsymbol{\Phi}_{1|0}\\
    \vdots\\
    \mathbf{H}_{L_k}\boldsymbol{\Phi}_{k|0}\\
    \mathbf{H}_{G_k}\boldsymbol{\Phi}_{k|0}\\
    \end{bmatrix}
    \triangleq
    \begin{bmatrix}
    \mathbf{M}_{L_0}\\
    \mathbf{M}_{G_0}\\
    \mathbf{M}_{L_1}\\
    \mathbf{M}_{G_1}\\
    \vdots\\
    \mathbf{M}_{L_k}\\
    \mathbf{M}_{G_k}\\
    \end{bmatrix},
\end{aligned}
\end{equation}
where $\mathbf{H}_{L_0},\mathbf{H}_{G_0}$ are given by (\ref{eq:HLk}) and (\ref{eq:HGk}), and $k$ is replaced with $0$. For $\mathbf{M}_{L_i}$ and $\mathbf{M}_{G_i}, i=1 \cdots k$, we have,
\begin{equation}\nonumber
    \mathbf{M}_{L_i}=\mathbf{H}_{\pi_i}\begin{bmatrix}
     \mathbf{M}_{L}(1)&
     -^{I_i}\hat{\mathbf{R}}_{L}\Delta _i&-^{I_i}\hat{\mathbf{R}}_{L}&^{I_i}\hat{\mathbf{R}}_{L}&\mathbf{0}&\mathbf{0}\\
     \end{bmatrix},\\
\end{equation}
\begin{equation}
    \nonumber
    \begin{aligned}
      \mathbf{M}_{G_i}=&\mathbf{H}_{\pi_i}^{'}
      [
      \mathbf{M}_{G}(1)&-^{I_i}\hat{\mathbf{R}}_{L}\Delta _i\quad-{}^{I_i}\hat{\mathbf{R}}_{L}\quad\mathbf{0}\quad\mathbf{M}_{G}(2)\\&\mathbf{M}_{G}(3)
      ],
    \end{aligned}
\end{equation}
where 
\begin{equation}
    \nonumber
    \begin{aligned}
     \mathbf{M}_{L}(1) &= {}^{I_i}\hat{\mathbf{R}}_{L}(^{L}\hat{\mathbf{p}}_{f_i}-^{L}\hat{\mathbf{p}}_{I_0}-^{L}\hat{\mathbf{v}}_{I_0}\Delta _i+\frac{1}{2}\mathbf{g}\Delta _{i}^{2})_{\times}{}^{I_0}\hat{\mathbf{R}}_{L}^{\top},\\ 
     \mathbf{M}_{G}(1) &= {}^{I_i}\hat{\mathbf{R}}_{L}(^{G_i}\hat{\mathbf{R}}_{L}^{\top}(^{G}\mathbf{p}_{F}-{}^{G_i}\hat{\mathbf{p}}_{L})-{}^{L}\hat{\mathbf{p}}_{I_0}-{}^{L}\hat{\mathbf{v}}_{I_0}\Delta _i\\&+\frac{1}{2}\mathbf{g}\Delta _{i}^{2})_{\times}{}^{I_0}\hat{\mathbf{R}}_{L}^{\top},\\
     \mathbf{M}_{G}(2)&=-{}^{I_i}\hat{\mathbf{R}}_{L}{}^{G}\hat{\mathbf{R}}_{L}^{\top}(^{G}\mathbf{p}_{F}-^{G_i}\hat{\mathbf{p}}_{L})_{\times},\\
     \mathbf{M}_{G}(3)&=-^{I_i}\hat{\mathbf{R}}_{L}{}^{G_i}\hat{\mathbf{R}}_{L}^{\top}.\\
    \end{aligned}
\end{equation}

Note that in case (a), for each time step, the state value after propagation is the same as the state value after update, i.e., for the state elements whose differential equation equal to zero ($^{L}\mathbf{p}_{f},{}^{G}\mathbf{R}_{L},{}^{G}\mathbf{p}_{L}$), their values are always equal to the initial value. Therefore, in the observability matrix above, ${}^{G_0}\hat{\mathbf{R}}_{L}={}^{G_1}\hat{\mathbf{R}}_{L}={}^{G_k}\hat{\mathbf{R}}_{L}\triangleq {}^{G}\mathbf{R}_{L}$, so as ${}^{L}\mathbf{p}_{f},{}^{G}\mathbf{p}_{L}$.

From the above observability matrix, we can find that its null space would be
\begin{equation} \label{eq:null_space_norm}
    \text{Null}(\mathbf{M})=\begin{bmatrix}
    ^{I_0}\hat{\mathbf{R}}_{L}\mathbf{g}&\mathbf{0}_{3}\\
    (-^{L}\hat{\mathbf{v}}_{I_0})_{\times}\mathbf{g}&\mathbf{0}_{3}\\
    (-^{L}\hat{\mathbf{p}}_{I_0})_{\times}\mathbf{g}&\mathbf{I}_{3}\\
    (-^{L}\hat{\mathbf{p}}_{f_0})_{\times}\mathbf{g}&\mathbf{I}_{3}\\
    ^{G_0}\hat{\mathbf{R}}_{L}\mathbf{g}&\mathbf{0}_{3}\\
    \mathbf{0}_{3\times1}&-^{G}\mathbf{R}_{L}\\
    \end{bmatrix},
\end{equation}
whose dimension is four.
\textcolor{black}{
We can find that compared with the unobservable subspace of the visual-inertial system \cite{consistVIN}, the null space of the observability matrix of the augmented system (\ref{eq:null_space_norm}) has six more rows which are introduced due to the existence of the augmented variable ($^{G}\mathbf{T}_{L}$, the relative transformation between $L$ frame and $G$ frame). We will see in case (b), these six more lines are the root of the inconsistent problem of the augmented system.}

$\bullet$ For case (b):
In this situation, the predicted state value is usually not equal to the updated state value. Therefore, the first three columns of the transition matrix $\boldsymbol{\Phi}_{t_{k+1}|t_{0}}$ do not have the elegant form as (\ref{eq:prop_matrix}), so that the first column of (\ref{eq:null_space_norm}) will not be the null space of the observability matrix.

On the other hand, \textcolor{black}{as mentioned above, due to the introduction of the augmented variable $^{G}\mathbf{T}_{L}$, the null space of the observability matrix of the system is coupled with $^{G}\mathbf{R}_{L}$. If only local feature measurements are available, the estimated value of $^{G}\mathbf{R}_{L}$ will not be changed, and the unobservable subspace of the system will not be influenced by the $^{G}\mathbf{R}_{L}$. However, when the map-based measurements are available, the estimated value of $^{G}\mathbf{R}_{L}$ will be updated, which leads to} the estimated $^{G_0}\hat{\mathbf{R}}_{L}\neq{}^{G_1}\hat{\mathbf{R}}_{L}\neq{}^{G_k}\hat{\mathbf{R}}_{L}$, so the last three columns of (\ref{eq:null_space_norm}) would not be the null space of the observability matrix. Therefore, for the standard EKF, the unobservable dimension cannot be maintained, which will lead to inconsistency.

\section*{Acknowledgment}

This work was supported in part by the National \textcolor{black}{Natural} Science Foundation
of China under Grant 61903332 and Alibaba Group through Alibaba Innovative Research Program.

\ifCLASSOPTIONcaptionsoff
  \newpage
\fi

\begin{IEEEbiography}[{\includegraphics[width=1in,clip,keepaspectratio]{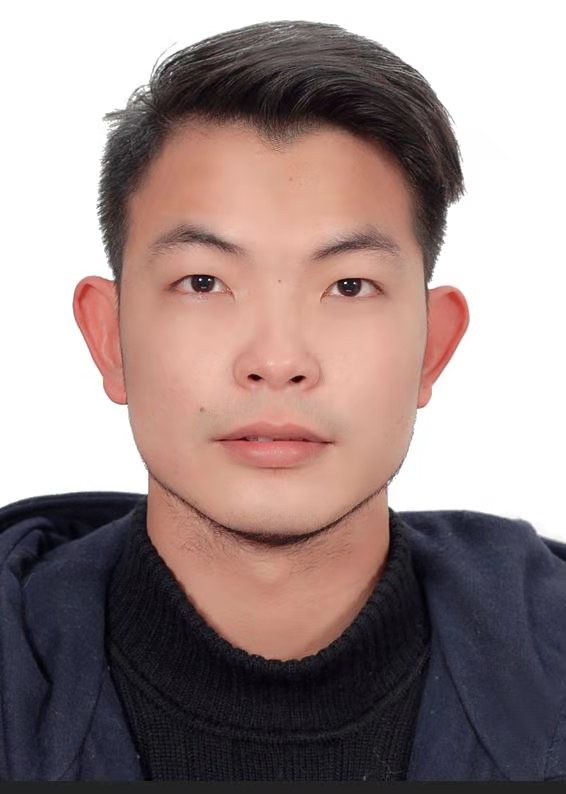}}]{Zhuqing Zhang} received his B.S from School of Astronautic, Northwestern Polytechnical University in 2017 and received his M.S. from School of Aeronautics and Astronautics, Shanghai Jiao Tong University in 2020. \\
\quad He is currently a Ph.D. candidate in the Department of Control Science and Engineering, Zhejiang University, Hangzhou, P.R. China. His lastest reasearch interest is multi sensor fusion localization.
\end{IEEEbiography}
\begin{IEEEbiography}[{\includegraphics[width=1in,clip,keepaspectratio]{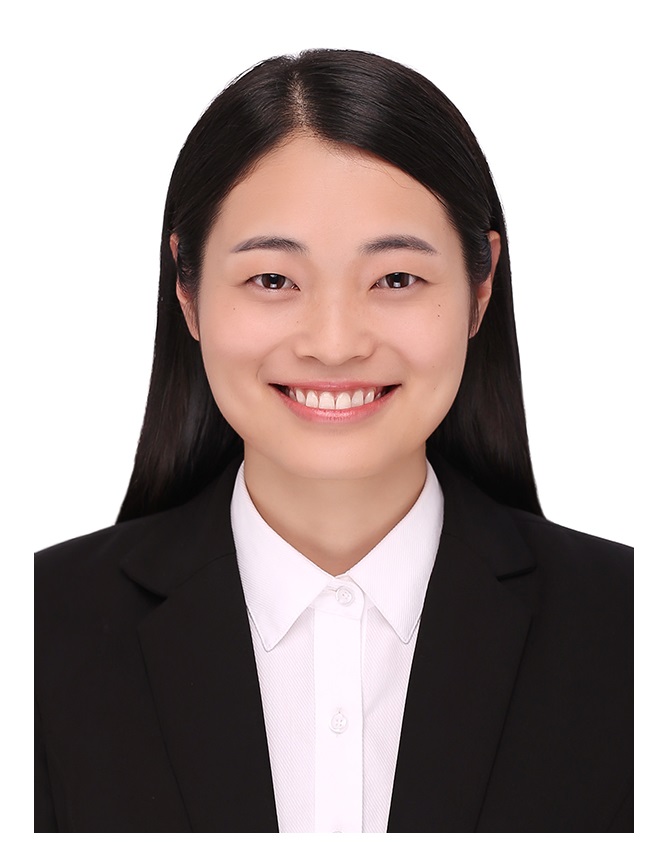}}]{Yanmei Jiao} received her BS in Intelligence Science and Technology from the Department of Computer Science and Control Engineering, Nankai University, Tianjin, P.R. China in 2017. \\
\quad She is currently a Ph.D. candidate in the Department of Control Science and Engineering, Zhejiang University, Hangzhou, P.R. China. Her research interests include computer vision and vision based localization.
\end{IEEEbiography}
\begin{IEEEbiography}[{\includegraphics[width=1in,clip,keepaspectratio]{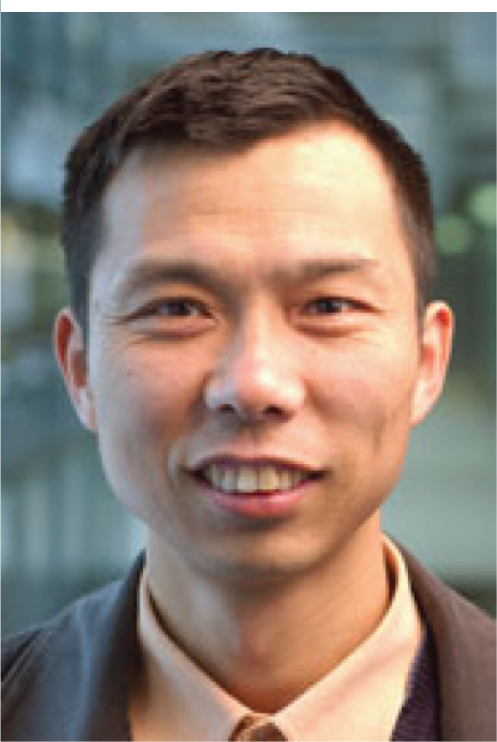}}]{Shoudong Huang} received the Bachelor and Master degrees in Mathematics, Ph.D. in Automatic Control from Northeastern University, PR China in 1987, 1990, and 1998, respectively. He is currently an Associate Professor at Centre for Autonomous Systems, Faculty of Engineering and Information Technology, University of Technology, Sydney, Australia. His research interests include nonlinear control systems and mobile robots simultaneous localization and mapping (SLAM), exploration and navigation.
\end{IEEEbiography}
\begin{IEEEbiography}[{\includegraphics[width=1in,clip,keepaspectratio]{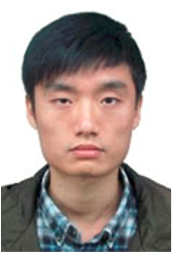}}]{Yue Wang} received his PhD in Control Science and Engineering from Department of Control Science and Engineering, Zhejiang University, Hangzhou, P.R. China in 2016. \\
\quad He is currently an Associate Professor in the Department of Control Science and Engineering, Zhejiang University, Hangzhou, P.R. China. His latest research interests include mobile robotics and robot perception.
\end{IEEEbiography}
\begin{IEEEbiography}[{\includegraphics[width=1in,clip,keepaspectratio]{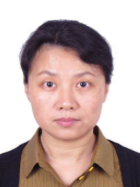}}]{Rong Xiong} received her PhD in Control Science and Engineering from the Department of Control Science and Engineering, Zhejiang University, Hangzhou, P.R. China in 2009. \\
\quad She is currently a professor in the Department of Control Science and Engineering, Zhejiang University, Hangzhou, P.R. China. Her latest research interests include motion planning and SLAM.
\end{IEEEbiography}




\end{document}